\documentclass[10pt,twocolumn]{IEEEtran}
\usepackage{cite}

\usepackage{amsmath,amssymb,amsfonts}
\usepackage{algorithmic}
\usepackage{graphicx}
\usepackage{textcomp}
\usepackage{subcaption}
\usepackage{hyperref}
\usepackage{bm}
\usepackage{multirow}
\usepackage{hhline}
\usepackage{extarrows}
\usepackage{mathtools}

\graphicspath{{./figures/}}
\newcommand{\R}{\mathbb{R}}
\newcommand{\N}{\mathbb{N}}

\newcommand{\Db}{\mathbb{D}}
\DeclareMathOperator*{\col}{col}

\newcommand{\svdots}{\raisebox{3pt}{$\scalebox{.75}{\vdots}$}} 
\newcommand{\sddots}{\raisebox{3pt}{$\scalebox{.75}{$\ddots$}$}} 

\newtheorem{theorem}{Theorem}

\newtheorem{remark}{Remark}
\newtheorem{definition}{Definition}
\newtheorem{example}{Example}
\newtheorem{assumption}{Assumption}
\newtheorem{proposition}{Proposition}

\newcommand{\RW}[1]{{#1}}

\newcommand{\Rev}[1]{{#1}}
\newcommand{\RevF}[1]{{#1}}
\newcommand{\HL}[1]{{#1}}

\title{Recurrent Equilibrium Networks:\\ Flexible Dynamic Models with Guaranteed Stability and Robustness}

\author{Max Revay$^\star$, Ruigang Wang$^\star$, Ian R. Manchester 
\thanks{$^\star$M. Revay and R. Wang made equal contribution to this paper. This work was supported by the Australian Research Council, grant DP190102963.}
\thanks{The authors are with the Australian Centre for Robotics and School of Aerospace, Mechanical and Mechatronic Engineering, The University of Sydney, Sydney, NSW 2006, Australia (e-mail: {\tt\small  ian.manchester@sydney.edu.au}).}%
}

\begin{document}

\maketitle

\begin{abstract}
This paper \HL{introduces \textit{recurrent equilibrium networks} (RENs), a new class of nonlinear dynamical models} for applications in machine learning, system identification and control. The new model class admits ``built in'' behavioural guarantees of stability and robustness. All models in the proposed class are contracting -- a strong form of nonlinear stability -- and models can satisfy prescribed incremental integral quadratic constraints (IQC), including Lipschitz bounds and incremental passivity. RENs are otherwise very flexible: they can represent all stable linear systems, all previously-known sets of contracting recurrent neural networks and echo state networks, all deep feedforward neural networks,  and all stable Wiener/Hammerstein models, and can approximate all fading-memory and contracting nonlinear systems. RENs are parameterized directly by a vector in $\mathbb{R}^N$, i.e. stability and robustness are ensured without parameter constraints, which simplifies learning since \HL{generic methods for unconstrained optimization such as stochastic gradient descent and its variants can be used}. The performance and robustness of the new model set is evaluated on benchmark nonlinear system identification problems, and the paper also presents applications in data-driven nonlinear observer design and control with stability guarantees.
\end{abstract}

\section{Introduction}
Deep neural networks (DNNs), recurrent neural networks (RNNs), and related models have revolutionised many fields of engineering and computer science \cite{lecunDeepLearning2015}. 
Their remarkable flexibility, accuracy, and scalability has led to renewed interest in neural networks in many domains including learning-based/data-driven methods in control, identification, and related areas (see e.g. \cite{levine2016end, yin2021imitation, brunke2022safe} and references therein). 

However, it has been observed that neural networks can \Rev{be} very sensitive to small changes in inputs \cite{szegedy2013intriguing}, and this sensitivity can extend to control policies \cite{russo2021towards}. Furthermore, their scale and complexity makes them difficult to certify for use in safety-critical systems, and it can be difficult to incorporate prior physical knowledge into a neural network model, e.g. that a model should be stable.
The most accurate current methods for certifying stability and robustness of DNNs and RNNs are based on mixed-integer programming \cite{tjengEvaluatingRobustnessNeural2018} and semidefinite programming \cite{raghunathanCertifiedDefensesAdversarial2018, fazlyab2019efficient} both of which face challenges when scaling to large networks.

In this paper, \HL{we introduce a new model structure: the \textit{recurrent equilibrium network} (REN)}.
\begin{enumerate}
	\item RENs are highly \textit{flexible} and include many established models as special cases, including DNNs, RNNs, echo-state networks and stable linear dynamical systems.
	\item RENs admit \textit{built in behavioural guarantees} such as stability, incremental gain, passivity, or other properties that are relevant to safety critical systems, \RevF{and are compatible with most existing frameworks for nonlinear/robust stability analysis}.
	\item RENs are \textit{easy to use} as they permit a direct (smooth, unconstrained) parameterization enabling learning of large-scale models \HL{via generic unconstrained optimization algorithms and off-the-shelf automatic-differentiation tools}. 
\end{enumerate}
\Rev{A REN is a dynamical model incorporating an \textit{equilibrium network} \cite{bai2019deep, winston2020monotone, revay2020lipschitz} , a.k.a. \textit{implicit network} \cite{ghaoui2019implicit}. Equilibrium networks are ``implicit depth'' neural networks, in which the output is generated as the zero set of an equation relating inputs and outputs, which can be viewed as the equilibrium of a ``fast'' dynamical system. This implicit structure brings the remarkable flexibility alluded to above, but also raises the question of existence and uniqueness of solutions, i.e. well-posedness. A benefit of our  parameterization approach is that the resulting RENs are always well-posed.}

RENs can be constructed to be contracting \cite{Lohmiller:1998}, a strong form of nonlinear stability, and/or to satisfy robustness guarantees in the form of  incremental integral quadratic constraints (IQCs) \cite{megretski1997system}. This class of constraints includes user-definable bounds on the network's Lipschitz constant (incremental gain), \Rev{which can be used to trade off performance vs sensitivity to adversarial perturbations}. The IQC framework also encompasses many commonly used tools for certifying stability and performance of system interconnections, including passivity methods in robotics \cite{hatanaka2015passivity}, networked-system analysis via dissipation inequalities \cite{arcak2016networks}, $\mu$ analysis \cite{zhou1996robust}, and standard tools for analysis of nonlinear control systems \cite{vanderschaftL2GainPassivityNonlinear2017}. 

\subsection{Learning and Identification of Stable Models}

The problem of learning dynamical systems with stability guarantees appears frequently in system identification. When learning models with feedback it is not uncommon for the model to be unstable even if the data-generating system is stable. For linear models, various methods have been proposed to guarantee stability via regularization and constrained optimization \cite{maciejowskiGuaranteedStabilitySubspace1995, van2001identification, lacy2003subspace, Nallasivam11, miller2013subspace}.  For nonlinear models, there has also been a substantial volume of research on stability guarantees, e.g. for polynomial models \cite{tobenkinConvexOptimizationIdentification2010, tobenkinConvexParameterizationsFidelity2017, umenbergerMaximumLikelihoodIdentification2018, umenbergerSpecializedInteriorPointAlgorithm2018}, Gaussian mixture models \cite{khansari-zadehLearningStableNonlinear2011a}, and recurrent neural networks \cite{miller2018stable, umenbergerConvexBoundsEquation2019, Kolter:2019, revay2020contracting, revay2020convex}. However, the problem is substantially more complex than the linear case due to the many possible nonlinear model structures and differing definitions of nonlinear stability.  Contraction is a strong form of nonlinear stability \cite{Lohmiller:1998} which is particularly well-suited to problems in learning and system identification since it guarantees stability of \textit{all} solutions of the model, irrespective of inputs or initial conditions. This is important in learning since the purpose of a model is usually to simulate responses to previously unseen inputs. The works \cite{tobenkinConvexOptimizationIdentification2010, tobenkinConvexParameterizationsFidelity2017, umenbergerMaximumLikelihoodIdentification2018, umenbergerSpecializedInteriorPointAlgorithm2018, miller2018stable, revay2020contracting, revay2020convex} are guaranteed to find contracting models.

\subsection{Lipschitz Bounds for Neural Network Robustness}

Model \textit{robustness} can be characterized in terms of sensitivity to small perturbations in the input. It has recently been shown that recurrent neural network models can be extremely fragile \cite{cheng2020seq2sick}, i.e. small changes to the input produce dramatic changes in the output.

Formally, sensitivity and robustness can be quantified via \textit{Lipschitz bounds} on the input-output mapping associated with the model. In machine learning, Lipschitz constants are used in the proofs of generalization bounds \cite{Bartlett:2017}, analysis of expressiveness \cite{Zhou:2019} and guarantees of robustness to adversarial attacks \cite{Huster:2018,Qian:2019}. There is also ample empirical evidence to suggest that Lipschitz regularity (and model stability, where applicable) improves generalization in machine learning \cite{gouk2021regularisation} and system identification \cite{revay2020contracting}. In reinforcement learning \cite{suttonReinforcementLearningIntroduction2018}, it has recently been found that the Lipschitz constant of policies has a strong effect on their robustness to adversarial attack \cite{russoOptimalAttacksReinforcement2019a}.  In \cite{kawanoDesignPrivacyPreservingDynamic2020} it was shown that privacy preservation in dynamic feedback policies can be represented as  an $\ell^2$ Lipschitz bound.

Unfortunately, even calculation of the Lipschitz constant of  feedforward (static) neural networks is NP-hard \cite{virmax2018lipschitz}. The tightest tractable bounds known to date use incremental quadratic constraints to construct a behavioural description of the neural network activation functions \Rev{\cite{fazlyab2020safety}}, but using these results in training is complicated by the fact that the constraints are not jointly convex in model parameters and constraint multipliers. In \cite{pauli2021training}, Lipschitz bounded feedforward models were trained using the Alternating Direction Method of Multipliers, and in \cite{revay2020contracting}, an a custom interior point solver were used. However, the requirements to satisfy linear matrix inequalities at each iteration make these methods difficult to scale. In \cite{wang2023direct}, the authors introduced a direct parameterization of feedforward neural networks satisfying the bounds of \Rev{\cite{fazlyab2020safety}}, using techniques related to the present paper.

\subsection{Applications of Contracting and Robust Models in Data-Driven Control and Estimation}

An ability to learn flexible dynamical models with contraction, robustness, and other behavioural constraints has many potential applications in control and related fields, some of which we explore in this paper.

 In robotics, passivity constraints are widely used to ensure stable interactions e.g. in teleoperation, vision-based control, and multi-robot control \cite{hatanaka2015passivity} and interaction with physical environments (e.g. \cite{ferragutiEnergyTankBasedInteractive2015, shahriariAdaptingContactsEnergy2017}). More generally, methods based on quadratic dissipativity and IQCs are a powerful tool for the design of complex interconnected cyber-physical systems \cite{megretski1997system, arcak2016networks}. Within these frameworks, the proposed REN architecture can be used to learn subsystems that specify prescribed or parameterized IQCs, and \RevF{which} therefore cannot destabilize the system when interconnected with other components.

A classical problem in control theory is observer design: to construct a dynamical system that estimates the internal (latent) state of another system from partial measurements. \RevF{A recent approach is to search for a contracting dynamical system that can reproduce true system trajectories} \cite{manchester2018contracting,yi2020reduced}. In Section \ref{sec:obs}, we formulate the observer design problem as a supervised learning problem over a set of contracting nonlinear systems, and demonstrate the approach on an unstable nonlinear reaction diffusion PDE.

In optimization of linear feedback controllers, the classical Youla-Kucera (or $Q$) parameterization provides a convex formulation for searching over all stabilizing controllers via a ``free'' stable \RevF{linear} system  parameter \cite{youlaModernWienerHopfDesign1976, zhou1996robust, hespanhaLinearSystemsTheory2018}. This approach can be extended to nonlinear systems \cite{fujimotoCharacterizationAllNonlinear2000a, vanderschaftL2GainPassivityNonlinear2017} in which the ``free parameter'' is a stable nonlinear model. In Sec. \ref{sec:youla}, \RevF{we} apply this idea to optimize nonlinear feedback policies for constrained linear control.

\subsection{Convex and Direct Parameterizations}\label{sec:intro_params}

\HL{The central contributions of this paper are new model parameterizations} which have behavioral constraints, and which are amenable to optimization. The first set of parameterizations we introduce includes (convex) linear matrix inequality (LMI) constraints, building upon \cite{tobenkinConvexOptimizationIdentification2010, revay2020convex}. LMI constraints can be incorporated into a learning process either through introduction of barrier functions or projections. However, they are computationally challenging for large-scale models. For example, a path-following interior point method, as proposed in \cite{revay2020convex} generally requires computing gradients of barrier functions, line search procedures, and a combination of ``inner'' and ``outer'' iterations as the barrier parameter changes.

To address this challenge, in this paper we also introduce \textit{direct} parameterizations of contracting and robust RENs. That is, we construct a smooth mapping from $\mathbb R^N$ to the model weights such that every model in the image of this mapping satisfies the desired behavioural constraints. This can be thought of as constructing a (redundant) intrinsic coordinate system on the constraint manifold. The construction is related to the method of \cite{burer2003nonlinear} for semidefinite programming, in which a positive-semidefinite matrix is parameterized by square-root factors. Our parameterization differs in that it avoids introducing any nonlinear equality constraints.

\HL{As mentioned above, direct parameterization allows generic optimization methods such as stochastic gradient descent (SGD) and ADAM \cite{Kingma:2017} to be applied.} Another advantage is that it allows easy \textit{random sampling} of nonlinear models with the required stability and robustness constraints by simply sampling a random vector in $\mathbb R^N$. This allows straightforward generation of \textit{echo state networks} with prescribed behavioral properties, i.e. large-scale recurrent networks with fixed dynamics and learnable output maps (see, e.g., \cite{buehnerTighterBoundEcho2006, yildizRevisitingEchoState2012} and references therein).

\subsection{Structure of this Paper}

The paper structure is as follows:
\begin{itemize}
  \item Sections \ref{sec:learning_stable} - \ref{sec:express} discuss the proposed model class and its properties. Section \ref{sec:learning_stable} formulates the problem of learning stable and robust dynamical models; in Section \ref{sec:ren} we present the REN model class; in Section \ref{sec:convex} we present convex parameterizations of stable and robust RENs; in Section \ref{sec:direct} we present direct (unconstrained) parameterisations of RENs; in Section \ref{sec:express} we discuss the expressivity of the REN model class, showing it includes many commonly-used models as special cases.
  \item Sections \ref{sec:ID} - \ref{sec:youla} present applications of learning stable/robust nonlinear models. Section \ref{sec:ID} presents applications to system identification; Section \ref{sec:obs} presents applications to nonlinear observer design; Section \ref{sec:youla} presents applications to nonlinear feedback design for linear systems. Associated Julia code is available in the package {\tt RobustNeuralNetworks.jl} \cite{barbara2023robustneuralnetworksjl}.
\end{itemize}
\Rev{A preliminary conference version was presented in \cite{revay2021recurrent}. The present paper expands the class of robustness properties to more general dissipativity conditions, removes the restriction that the model has zero direct-feedthrough, introduces the acyclic REN, adds proofs of all theoretical results, adds new material on echo state networks, and includes novel approaches to nonlinear observer design and optimization of feedback controllers enabled by the REN.}

\subsection{Notation}
The set of sequences $ x:\N\rightarrow\R^n $ is denoted by $ \ell_{2e}^n $. Superscript $ n $ is omitted when it is clear from the context.  For $x\in\ell_{2e}^n$, $x_t\in\R^n$ is the value of the sequence $ x $ at time $t\in \N$. The subset $ \ell_2\subset \ell_{2e} $ consists of all square-summable sequences, i.e., $ x\in\ell_2 $ if and only if the $ \ell_2 $ norm $ \|x\|:=\sqrt{\sum_{t=0}^{\infty}|x_t|^2} $ is finite, where $ |(\cdot)|$ denotes Euclidean norm. Given a sequence $ x\in\ell_{2e} $, the $ \ell_2 $ norm of its truncation over $ [0,T] $ is $ \|x\|_T:=\sqrt{\sum_{t=0}^{T}|x_t|^2} $. For two sequences $x,y\in \ell_{2e}^n$, the inner product over $[0,T]$ is $\langle x,y\rangle_T:=\sum_{t=0}^{T}x_t^\top y_t$.
We use $A\succ 0$ and $A\succeq0$ to denote a positive definite and positive semi-definite matrix, respectively.
We denote the set of positive-definite diagonal matrices by $\mathbb{D}_+$. Given a positive-definite matrix $P$ we use $|\cdot|_P$ to denote the weighted Euclidean norm, i.e. $|a|_P = \sqrt{a^\top Pa}$.

\section{Learning Stable and Robust Models}
\label{sec:learning_stable}
This paper is concerned with \textit{learning} of nonlinear dynamical models, i.e. finding a particular model within a set of candidates using some \textit{data} relevant to the problem at hand. \HL{The central aim of this paper is to construct model classes} that are \textit{flexible} enough to make full use of available data, and yet \textit{guaranteed} to be well-behaved in some sense.

Given a dataset $\tilde{z}$, we consider the problem of learning a nonlinear state-space dynamical model of the form
\begin{equation}\label{eq:rnn}
	x_{t+1}=f(x_t,u_t,\theta), \quad y_t=g(x_t,u_t,\theta) 
\end{equation}
that minimizes some loss or cost function depending (in part) on the data, i.e. to solve a problem of the form
\begin{equation}\label{eq:learning}
	\min_{\theta\in \Theta} \; \mathcal{L}(\tilde{z},\theta).
\end{equation}
In the above, $x_t\in\R^n,u_t\in\R^m,y_t\in\R^p,\theta\in\Theta\subseteq\R^N$ are the model state, input, output and parameters, respectively. Here $ f:\R^n\times\R^m\times\Theta\rightarrow\R^n$ and $g:\R^n\times\R^m\times\Theta\rightarrow\R^p$ are piecewise continuously differentiable functions.

\begin{example}
	In the context of system identification we may have $\tilde z = (\tilde y, \tilde u)$ consisting of finite sequences of input-output measurements, and aim to minimize {\em simulation error}: 
\begin{equation}\label{eq:sim_error}
 \mathcal{L}(\tilde{z},\theta)=\|y-\tilde{y}\|_T^2 
\end{equation}
where $ y=\mathfrak{R}_a(\tilde{u}) $ is the output sequence generated by the nonlinear dynamical model \eqref{eq:rnn} with initial condition $x_0=a$ and inputs $u_t=\tilde{u}_t$. Here the initial condition $a$ may be part of the data $\tilde z$, or considered a learnable parameter in $\theta$.
\end{example}

\HL{The main contributions of this paper are model parameterizations}, and we make the following definitions:

\begin{definition}\label{def:direct} 
	A model parameterization \eqref{eq:rnn} is called a \emph{convex parameterization} if $\Theta\subseteq \R^{N}$ is a convex set. Furthermore, it is called a \emph{direct parameterization} if $\Theta=\R^N$.
\end{definition}

Direct parameterizations are useful for learning large-scale models since many scalable unconstrained optimization methods (e.g. stochastic gradient descent) can be applied to solve  \eqref{eq:learning}. We will parameterize stable nonlinear models, and the particular form of stability we use is the following:
\begin{definition}\label{def:contracting}
	A model \eqref{eq:rnn} is said to be \textit{contracting with rate $\alpha\in (0,1)$} if for any two initial conditions $ a, b\in\R^n $, given the same input sequence $u\in \ell_{2e}^m$, the state sequences $ x^a $ and $ x^b $ satisfy \begin{equation}\label{eq:contraction} |x_t^a-x_t^b|\leq K\alpha^t|a-b| \end{equation} for some $K>0$. 
\end{definition}

Roughly speaking, contracting models forget their initial conditions exponentially. Beyond stability, we will also consider robustness constraints of the following form:

\begin{definition}
	A model \eqref{eq:rnn} is said to satisfy the \emph{incremental integral quadratic constraint} (IQC) defined by $(Q,S,R)$ where $0\succeq Q\in \R^{p\times p}$, $S\in \R^{m\times p}$, and $R=R^\top\in \R^{m\times m}$, if for all pairs of solutions with initial conditions $ a,b\in\R^n $ and input sequences $ u,v\in \ell_{2e}^m $, the output sequences $ y^a=\mathfrak{R}_a(u) $ and $y^b=\mathfrak{R}_b(v)$ satisfy
	\begin{equation}\label{eq:iqc-def}
		\sum_{t=0}^{T}\begin{bmatrix}
		y_t^a - y_t^b \\ u_t-v_t
		\end{bmatrix}^\top
		\begin{bmatrix}
		Q & S^\top \\ S & R
		\end{bmatrix}
		\begin{bmatrix}
		y_t^a - y_t^b \\ u_t-v_t
		\end{bmatrix}\geq -d(a,b),\; \forall T
	\end{equation} 
	for some function $d(a,b) \geq0$ with $d(a,a) = 0$.
\end{definition}

Important special cases of incremental IQCs include:
\begin{itemize}
\item $Q=-\frac{1}{\gamma}I, R=\gamma I, S=0$: the model satisfies an $\ell^2$ Lipschitz bound, a.k.a. incremental $\ell^2$-gain bound, of $\gamma$:
\begin{equation*}
	\|\mathfrak{R}_a(u)-\mathfrak{R}_a(v)\|_T\leq \gamma \|u-v\|_T,\;.
\end{equation*} 
for all $u,v\in\ell_{2e}^m,\, T\in \N.$
\item $Q = 0, R = -2\nu I, S=I$ where $\nu\ge 0$: the model is  monotone on $\ell^2$ (strongly if $\nu >0$), a.k.a. incrementally passive (incrementally strictly input passive, resp.):
\[
\langle \mathfrak{R}_a(u) - \mathfrak{R}_a(v), u – v \rangle_T \ge \nu \|u-v\|_T^2
\]
for all $ u,v\in\ell_{2e}^m$ and $ T\in \N$. 

\item $Q = -2\rho I, R = 0, S=I$ where $\rho >0$: the model is incrementally strictly output passive:
\[
    \langle \mathfrak{R}_a(u) - \mathfrak{R}_a(v), u – v \rangle_T \ge \rho\|\mathfrak{R}_a(u)-\mathfrak{R}_a(v)\|_T^2
\]
for all $ u,v\in\ell_{2e}^m$ and $T\in \N$. If $\rho=1$ the model is \textit{firmly nonexpansive} on $\ell^2$.
\end{itemize}
In other contexts, $Q, S, R$ may themselves be decision variables in a separate optimization problem to ensure stability of interconnected systems (see, e.g., \cite{megretski1997system,arcak2016networks} .

\begin{remark}
	 Given a model class guaranteeing incremental IQC defined by constant matrices $Q,S,R$, it is straightforward to construct models satisfying \textit{frequency-weighted} IQCs. E.g. by constructing a model $\mathfrak R$ that is contracting and satisfies an $\ell^2$ Lipschitz bound, and choosing stable linear filters $\bm{W}_1, \bm{W}_2$, with $\bm{W}_1$ having a stable inverse, the new model
\[
y = \mathfrak{W}_a(u) = \bm{W}_1^{-1}\mathfrak R_a(\bm{W}_2 u)
\]
is contracting and satisfies the frequency-weighted bound
\[
	\|\bm{W}_1(\mathfrak{W}_a(u)-\mathfrak{W}_a(v))\|_T\leq \gamma \|\bm{W}_2(u-v)\|_T.
\]
\end{remark}

\section{Recurrent Equilibrium Networks}
\label{sec:ren}
\HL{The model structure we propose -- the \emph{recurrent equilibrium network} (REN)} -- is a state-space model of the form \eqref{eq:rnn} with
\begin{align}
x_{t+1} &= Ax_t+B_1w_t+B_2u_t+b_x,\label{eq:ren-state}\\  y_t &= C_{2}x_t+D_{21}w_t+D_{22}u_t+b_y, \label{eq:ren-output}
\end{align}
in which $w_t$ is the solution of an \textit{equilibrium network}, a.k.a. \textit{implicit network} \cite{bai2019deep,winston2020monotone,revay2020lipschitz, ghaoui2019implicit}:
\begin{equation}\label{eq:implicit}
w_t=\sigma(D_{11} w_t+C_1x_t+D_{12}u_t+b_v),
\end{equation}
\RevF{where $A,B_\cdot, C_\cdot, D_\cdot$ are matricies of appropriate dimension, $b_x\in\R^n,b_y\in\R^p,b_v\in\R^q$ are ``bias'' vectors, }and $\sigma$ is a scalar nonlinearity applied elementwise, referred to as an ``activation function''. We will show below how to ensure that a unique solution $w_t^*$ to \eqref{eq:implicit} exists and can be computed efficiently. 

\begin{remark} The term ``equilibrium'' comes from the fact that any solution of the above implicit equation is also an equilibrium point of the difference equation $w_t^{k+1}=\sigma(Dw_t^k+b_w)$ or the ordinary differential equation $\frac{d}{ds}w_t(s)=-w_t(s)+\sigma(Dw_t(s)+b_w) $, where $b_w=C_1x_t+D_{12}u_t+b_v$ is considered ``frozen'' for each $t$. One interpretation of the REN model is that it represents a two-timescale or singular perturbation model, in which the ``fast'' dynamics in $w$ are assumed to reach the equilibrium \eqref{eq:implicit} well within each time-step of the ``slow'' dynamics in $x$ \eqref{eq:ren-state}.	
\end{remark}

\begin{figure}[!bt]
	\centering
	\includegraphics[width=0.6\linewidth, trim={0cm, 0cm, 0.0cm, 0cm}, clip]{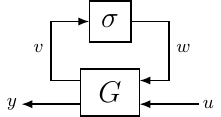}
	\caption{\label{fig:feedback} REN as a feedback interconnection of a linear system $G$ and a nonlinear activation $\sigma$.}
\end{figure}

It will be convenient to represent the REN model as a feedback interconnection of a linear system $G$ and a memoryless nonlinear operator $\sigma$, as depicted in Fig. \ref{fig:feedback}:
\begin{align}
\renewcommand\arraystretch{1.2}
\begin{bmatrix}
x_{t+1} \\ v_t \\ y_t
\end{bmatrix}&=
\overset{W}{\overbrace{
		\left[
		\begin{array}{c|cc}
		A & B_1 & B_2 \\ \hline 
		C_{1} & D_{11} & D_{12} \\
		C_{2} & D_{21} & D_{22}
		\end{array} 
		\right]
}}
\begin{bmatrix}
x_t \\ w_t \\ u_t
\end{bmatrix}+
\overset{b}{\overbrace{
		\begin{bmatrix}
		b_x \\ b_v \\ b_y
		\end{bmatrix}
}}, \label{eq:G}\\
w_t=\sigma(&v_t):=\begin{bmatrix}
\sigma(v_{t}^1) & \sigma(v_{t}^2) & \cdots & \sigma(v_{t}^q)
\end{bmatrix}^\top, \label{eq:sigma}
\end{align}
where $v_t,w_t\in\R^q$ are the input and output of activation functions respectively. The learnable parameter is $\theta:=\{W, b\}$ where $W\in \R^{(n+q+p)\times(n+q+m)}$ is the weight matrix, and $b\in \R^{n+q+p}$ the bias vector. Typically the activation function $\sigma$ is fixed, although this is not essential.

\subsection{Flexibility of Equilibrium Networks}\label{sec:feedforward}

\Rev{In \cite{revay2020convex} we introduced and studied a class of models similar to \eqref{eq:ren-state}, \eqref{eq:ren-output}, \eqref{eq:implicit} with the exception that $D_{11}$ was absent\footnote{Note that \cite{revay2020convex} used different notation, so in that paper it was actually $D_{22}$ which was absent, corresponding to $D_{11}$ in the notation of the present paper.}. This apparently minor change to the model has far-reaching consequences in terms of greatly increased representational flexibility and significantly simpler learning algorithms, while also requiring assurances about existence of solutions and their efficient computation.

With $D_{11}=0$, the network \eqref{eq:implicit} is simply a single-layer neural network.} In contrast, equilibrium networks ($D_{11}\ne 0$) are much more flexible, with many commonly-used feedforward network architectures included as special cases. For example, consider a standard $L$-layer deep neural network:
\begin{align}
z_0&=u,\notag\\ z_{l+1}&=\sigma(W_l z_l+b_l),\quad l=0,...,L-1 \label{eq:ff_network}\\ 
y&=W_Lz_L+b_L \notag
\end{align}
where $z_l$ is the output of the $l$th hidden layer. This can be written as an equilibrium network with
\begin{gather*} 
w=\col(z_1,\ldots,z_L),\quad  b_v=\col(b_0,\ldots,b_{L-1}), \quad b_y=b_L\\
\Rev{C_1=0,\quad  C_2=0,\quad D_{21}=\begin{bmatrix}
0 & \cdots & 0 & W_L
\end{bmatrix},\quad D_{22}=0 }, \\
D_{11}=\begin{bmatrix}
0 & & &\\
W_1 & \sddots & & \\
\svdots & \sddots & 0 & \\
0 & \cdots & W_{L-1} & 0 
\end{bmatrix},\quad 
D_{12}=
\begin{bmatrix}
    W_0 \\ 0 \\ \svdots \\ 0
\end{bmatrix}.
\end{gather*}
Equilibrium networks can represent many other interesting structures including residual, convolution, and other feedforward networks. The reader is referred to \cite{bai2019deep, ghaoui2019implicit, winston2020monotone, revay2020lipschitz} for further discussion of equilibrium networks and their properties.

Allowing $D_{11}$ to be non-zero is also key to our construction of {direct} paramaterizations of contracting and robust RENs (in Sec. \ref{sec:direct}). \HL{As discussed in Section \ref{sec:intro_params} this enables model learning via simple and generic first-order optimization methods, whereas \cite{revay2020convex} required a specialized interior-point method to deal with model behavioural constraints.} Direct parameterization also enables easy random sampling of contracting models, so-called \textit{echo state networks} (see Sec. \ref{sec:echostate}) and  this enables convex learning of nonlinear feedback controllers (see Sec. \ref{sec:youla}).

\subsection{Well-posedness of Equilibrium Networks and Acyclic RENs}
The added flexibility of equilibrium networks comes at a price: depending on the value of $D_{11}$, the implicit equation \eqref{eq:implicit} may or may not admit a unique solution $w_t$ for a given $x_t, u_t$. An equilibrium network or REN is \textit{well-posed} if a unique solution is guaranteed. In  \cite{revay2020lipschitz} it was shown that if there exists a $\Lambda\in \Db_+^{n}$ such that 
\begin{equation}\label{eq:W}
2\Lambda-\Lambda \Rev{D_{11}}-\Rev{D_{11}}^\top \Lambda \succ 0,
\end{equation}
then the equilibrium network is well-posed. We will show in Theorem \ref{prop:1} below that this is always satisfied for our proposed model parameterizations.

A useful subclass of REN that is trivially well-posed is the \emph{acyclic REN} where the weight $D_{11}$ is constrained to be strictly lower triangular. In this case, the elements of $w_t$ can be explicitly computed row-by-row from \eqref{eq:implicit}. We can interpret $D_{11}$ as the adjacency matrix of a directed graph defining interconnections between the neurons  in the equilibrium network and if $D_{11}$ is strictly lower triangular then this graph is guaranteed to be acyclic. Compared to the general REN, the acyclic REN is simpler to implement and in our experience often provides models of similar quality, as will be discussed in Sec. \ref{sec:sysid results}.

\subsection{Evaluating RENs and their gradients}

For a well-posed REN with full $D_{11}$, solutions can be computed by formulating an equivalent monotone operator splitting problem \cite{ryu2016primer}. In the authors' experience, the Peaceman Rachford algorithm is reliable and efficient \cite{revay2020lipschitz}.

When training an equilibrium network via gradient descent, we need to compute the Jacobian $\partial w_t^*/\partial (\cdot)$ where $w_t^*$ is the solution of the implicit equation \eqref{eq:implicit}, and $(\cdot)$ denotes the input to the network or model parameters. By using the implicit function theorem, $\partial w_t^*/\partial (\cdot)$ can be computed via   
\begin{equation}\label{eq:gradient}
  \frac{\partial w_t^*}{\partial (\cdot)}=(I-JD)^{-1}J\frac{\partial (Dw_t^\star+b_w)}{\partial (\cdot)}
\end{equation}
where $J$ is the Clarke generalized Jacobian of $\sigma$ at $Dw_t^*+b_w$. From Assumption~\ref{asmp:1} in Section~\ref{sec:cr-ren}, we have that $J$ is a singleton almost everywhere. 
It was shown in \cite{revay2020lipschitz} that Condition~\eqref{eq:W} implies matrix $I-J D$ is invertible.

 \subsection{Contracting and Robust RENs}\label{sec:cr-ren}

We call the model of \eqref{eq:G}, \eqref{eq:sigma} a contracting REN (C-REN) if it is contracting and a robust REN (R-REN) if it satisfies the incremental IQC. 
We make the following assumption on $\sigma$, which holds for commonly-used activation functions \cite{goodfellow2016deep}:
\begin{assumption}\label{asmp:1}
	The activation function $\sigma$ is piecewise differentiable and slope-restricted in $[0,1]$, i.e.,
	\begin{equation}\label{eq:slope}
	0\leq \frac{\sigma(y)-\sigma(x)}{y-x}\leq 1, \quad \forall x,y\in\R,\; x\neq y.
	\end{equation}
\end{assumption}
\Rev{The following theorem gives conditions for contracting and robust RENs: 
\begin{theorem}\label{prop:1}
	Consider the REN model \eqref{eq:G}, \eqref{eq:sigma} satisfying Assumption \ref{asmp:1}, and a given $\bar\alpha \in (0,1]$. 
\begin{enumerate}
  \item \textbf{Contracting REN:} suppose there exists $P=P^\top\succ 0$ and $\Lambda\in \Db_+$ such that
	\begin{equation}\label{eq:lmi-stable-explicit}
		\begin{bmatrix}
			\bar\alpha^2 P & -C_1^\top \Lambda \\
			-\Lambda C_1 & W
		\end{bmatrix}-
		\begin{bmatrix}
			A^\top \\ B_1^\top
		\end{bmatrix}P
		\begin{bmatrix}
			A^\top \\ B_1^\top
		\end{bmatrix}^\top\succ 0
	\end{equation}
	where $W=2\Lambda -\Lambda D_{11}-D_{11}^\top \Lambda$. Then the REN is well-posed and contracting with some rate $\alpha<\bar\alpha$.
	\item \textbf{Robust REN:} consider the incremental defined in IQC \eqref{eq:iqc-def} with given $(Q,S,R)$ where $Q\preceq 0$. Suppose there exist $P=P^\top\succ 0$ and $\Lambda\in \Db_+$ such that 
	\begin{equation}\label{eq:lmi-qsr-explicit}
		\begin{split}
			\begin{bmatrix}
				\bar\alpha^2 P & -C_1^\top \Lambda & C_2^\top S^\top\\
				-\Lambda C_1 & W & D_{21}^\top S^\top - \Lambda D_{12} \\
				S C_2 & S D_{21} - D_{12}^\top \Lambda & R + SD_{22}+D_{22}^\top S^\top
			\end{bmatrix} \\
			-\begin{bmatrix}
				A^\top \\ B_1^\top \\ B_2^\top 
			\end{bmatrix}P
			\begin{bmatrix}
				A^\top \\ B_1^\top \\ B_2^\top 
			\end{bmatrix}^\top +
			\begin{bmatrix}
				C_2^\top \\ D_{21}^\top \\ D_{22}^\top
			\end{bmatrix}Q
			\begin{bmatrix}
				C_2^\top \\ D_{21}^\top \\ D_{22}^\top
			\end{bmatrix}^\top \succ 0.
		\end{split}
	\end{equation}
 Then the REN is well-posed, satisfies \eqref{eq:iqc-def} and is contracting with a rate $\alpha<\bar\alpha$.
\end{enumerate}
\end{theorem}}
The proof can be found in Appendix~\ref{sec:proof-prop-1}. 
\Rev{
The main idea behind the LMI  for the contracting REN is to use an incremental Lyapunov function $V(\Delta x) = |\Delta x|^2_P$, where $\Delta x$ denotes the difference between a pair of solutions, and show that	
\begin{equation}\label{eq:lmi idea}
    V(\Delta x_{t+1}) \leq ~\alpha^2V(\Delta x_t)-\Gamma(\Delta v_t, \Delta w_t)
\end{equation}
and that $\Gamma(\Delta v_t, \Delta w_t)\ge 0$ for the activation function $\sigma$, where $\Gamma$ is an incremental quadratic constraint as in \cite{fazlyab2019efficient, revay2020convex} with a multiplier matrix $\Lambda$. The construction for the Robust REN is similar, but uses an incremental dissipation inequality.
}

\begin{remark}
Note that \eqref{eq:lmi-stable-explicit} and \eqref{eq:lmi-qsr-explicit} immediately imply that $W\succ 0$, which is precisely the equilibrium network well-posedness condition \eqref{eq:W}.
\end{remark}

\begin{remark}
	For a fixed REN model, Conditions \eqref{eq:lmi-stable-explicit} and \eqref{eq:lmi-qsr-explicit} are convex in the stability/performance certificate $P$ and IQC multiplier $\Lambda$. However they are not \textit{jointly}  convex in the model parameters $\theta$, certificate $P$, and multiplier $\Lambda$. We will resolve this in the next section.
\end{remark}

\begin{remark}
	The proof is based on IQC characterization of \eqref{eq:slope} with a diagonal multiplier matrix $\Lambda$. If signal boundedness is of interest rather than contraction, then one can use a richer class of multipliers designed for repeated nonlinearities \cite{chu1999bounds,damato2001new,kulkarni2002all}. However, these multipliers are not valid for incremental IQCs and contraction \cite{revay2020lipschitz}.
\end{remark}

\Rev{While $Q, S, R$ can be chosen so that a robust REN verifies a \textit{particular} Lipschitz bound $\gamma$, the following weaker property is true of contracting RENs:
\begin{theorem}\label{thm:2}
	Every contracting REN -- i.e. a model \eqref{eq:G}, \eqref{eq:sigma} satisfying Assumption \ref{asmp:1} and \eqref{eq:lmi-stable-explicit} -- satisfies the $\ell^2$ Lipschitz condition for some bound $\gamma<\infty$.
\end{theorem}
The proof is in Appendix \ref{sec:proof-thm-2}.}

\section{Convex Parameterizations of RENs}
\label{sec:convex}
In this section we propose convex parameterizations for C-RENs/R-RENs, which are based on the following implicit representation of the linear component $G$:
\begin{equation}\label{eq:G-implicit}
	\begin{bmatrix}
	E x_{t+1} \\ \Lambda v_t \\ y_t
	\end{bmatrix}= 
	\overset{\widetilde{W}}{\overbrace{
	\left[
	\begin{array}{ccc}
	F & \mathcal{B}_1 & \mathcal{B}_2 \\ 
	\mathcal{C}_{1} & \mathcal{D}_{11} & \mathcal{D}_{12} \\
	C_{2} & D_{21} & D_{22}
	\end{array} 
	\right]
	}}
	\begin{bmatrix}
	x_t \\ w_t \\ u_t
	\end{bmatrix}+\tilde{b}
\end{equation} 
where $E$ is an invertible matrix and $\Lambda$ is a positive-definite diagonal matrix. \Rev{The model parameters are $\theta_{\mathrm{cvx}}:=\{E,\Lambda, \widetilde{W}, \tilde{b}\}$. 

Note that $\theta_{\mathrm{cvx}}$ can easily be mapped to $\theta$ by multiplying the first and second rows of \eqref{eq:G-implicit} by $E^{-1}$ and $\Lambda^{-1}$, respectively.} 
Therefore the parameters $E$ and $\Lambda$ do not expand the model set, however the extra degrees of freedom will allow us to formulate sets of C-RENs and R-RENs that are jointly convex in the model parameters, stability certificate, and multipliers. 

\begin{definition}
	A model of the form \eqref{eq:G-implicit}, \eqref{eq:sigma} is said to be well-posed if it yields a unique $(w_t,x_{t+1})$ for any $x_t,u_t$ and $\tilde{b}$, and hence a unique response to any initial conditions and input.
\end{definition}

To construct a convex parameterization of C-RENs, we introduce the following LMI constraint:
\begin{equation}\label{eq:lmi-stable}
\Rev{H(\theta_{\mathrm{cvx}}):=}	\begin{bmatrix}
	E+E^\top-\frac{1}{\bar\alpha^2}\mathcal{P} & -\mathcal{C}_1^\top & F^\top \\
	-\mathcal{C}_1 & \mathcal{W}  & \mathcal{B}_1^\top \\
	F & \mathcal{B}_1 & \mathcal{P}
	\end{bmatrix} \succ 0,
\end{equation}
where $\mathcal{W}=2\Lambda -\mathcal{D}_{11}-\mathcal{D}_{11}^\top$. The convex parameterization of C-RENs is then given by 
\[
	\Theta_C:=\{\theta_{\mathrm{cvx}}\mid \exists\, \mathcal{P} = \mathcal{P}^\top\succ 0 \; \mathrm{s.t.}\; H(\theta_{\mathrm{cvx}})\succ 0\}.
\]
To construct convex parameterization of R-RENs, we propose the following convex constraint:
\begin{equation}\label{eq:lmi-qsr}
	\begin{split}
		& \begin{bmatrix}
			E+E^\top-\frac{1}{\bar\alpha^2}\mathcal{P} & -\mathcal{C}_1^\top & C_2^\top S^\top \\
			-\mathcal{C}_1 & \mathcal{W} & D_{21}^\top S^\top -\mathcal{D}_{12} \\
			S C_2 & SD_{21}-\mathcal{D}_{12}^\top & R+SD_{22}+D_{22}^\top S^\top
		\end{bmatrix} \\
		&-\begin{bmatrix}
			F^\top \\ \mathcal{B}_1^\top \\ \mathcal{B}_2^\top
		\end{bmatrix} \mathcal{P}^{-1}
		\begin{bmatrix}
			F^\top \\ \mathcal{B}_1^\top \\ \mathcal{B}_2^\top
		\end{bmatrix}^\top +
		\begin{bmatrix}
			C_2^\top \\ D_{21}^\top \\ D_{22}^\top
		\end{bmatrix}Q
		\begin{bmatrix}
			C_2^\top \\ D_{21}^\top \\ D_{22}^\top
		\end{bmatrix}^\top \succ 0
	\end{split}
\end{equation}
where $Q\preceq 0$, $S$, and $R$ are given. The convex parameterization of R-RENs is then defined as
\[
	\Theta_R:=\{\theta_{\mathrm{cvx}}\mid \exists \mathcal{P} = \mathcal{P}^\top\succ 0\; \mathrm{s.t.}\; \eqref{eq:lmi-qsr}\}.
\]

The following results relates the above parameterizations to the desired model behavioural properties:
\begin{theorem}\label{thm:convex_param}
	All models in $\Theta_C$ are well-posed and contracting with rate $\alpha<\bar\alpha$. All models in $\Theta_R$ are well-posed, contracting with rate $\alpha<\bar\alpha$, and satisfy the IQC defined by $(Q,S,R)$.
\end{theorem}
The proof can be found in the Appendix \ref{sec:proof-thm-1}.



\begin{remark}
With the convex parameterizations, is straightforward to enforce  any desired sparsity structure on $D_{11}$, e.g. corresponding to a multi-layer neural network as per Section \ref{sec:feedforward}. Since $\Lambda$ is diagonal, the sparsity structures of $\mathcal{D}_{11}$ and $D_{11}=\Lambda^{-1}\mathcal{D}_{11}$ are identical, and so the desired structure can be added as a linear constraint on $\mathcal{D}_{11}$.
\end{remark}


\section{Direct Parameterizations of RENs}
\label{sec:direct}
In the previous section we gave convex parameterizations of contracting and robust RENs in terms of linear matrix inequalities (LMIs), i.e.  intersections of the cone of positive semidefinite matrices with affine constraints. While convexity of a model set is useful, LMIs are challenging to verify for large-scale models, and especially to enforce during training.

\HL{In this section we provide direct parameterizations, i.e. smooth mappings from $\mathbb R^N$ to the weights and biases of a REN, enabling \textit{unconstrained} optimization methods to be applied.}  We do so by first constructing representations of RENs directly in terms of the positive semidefinite cone \textit{without} affine constraints, and then parameterize this cone in terms of its square-root factors.

\subsection{Direct Parameterizations of Contracting RENs}
\label{sec:direct_caren}

\Rev{
The key observation leading to our construction is that the mapping from contracting REN parameters $\theta_{\mathrm{cvx}}$ to $H$ in \eqref{eq:H} is \textit{surjective}, i.e. it maps onto the entire cone of positive-definite matrices. Furthermore, as we will show below it is straightforward to construct a (non-unique) inverse that maps from \textit{any} positive-definite matrix back to $\theta_{\mathrm{cvx}}$ defining a well-posed and contracting REN.

\paragraph{Free parameters} of the parameters in $\theta_{\mathrm{cvx}}$, the following have no effect on stability and can be freely parameterized in terms of their elements: $\mathcal{B}_2\in \R^{n\times m}$, $C_2\in \R^{p\times n}$, $\mathcal{D}_{12}\in\R^{q\times m}$, $ D_{21}\in \R^{p\times q}$, $ D_{22}\in \R^{p\times m}, \tilde b \in\R^{(2n+q)}$.


\paragraph{Constrained parameters, acyclic case}
the parameters $E,  F, \Lambda, \mathcal{B}_1$ and $\mathcal C_1$ relate to internal dynamics and therefore affect the stability properties of a REN. Here we construct them from two free matrix variables $ X\in \R^{(2n+q)\times (2n+q)}$ and $Y_1\in \R^{n\times n}$.
}

We first construct $H$ from $X$ as
\begin{equation}\label{eq:H}
	H=\begin{bmatrix}
		H_{11} & H_{12} & H_{13} \\
		H_{21} & H_{22} & H_{23} \\
		H_{31} & H_{32} & H_{33} 
	\end{bmatrix}=X^\top X+\epsilon I \succ 0
\end{equation}
where $\epsilon$ is a small positive scalar, and we have partitioned $H$ into blocks of size $n, n$, and $q$. Comparing \eqref{eq:H} to \eqref{eq:lmi-stable} we can immediately construct
\begin{equation}\label{eq:H_others}
	F = H_{31}, \quad
  \mathcal{B}_1 = H_{32}, \quad
  \mathcal{P} = H_{33},\quad
  \mathcal{C}_1 = -H_{21}.
\end{equation}
Further, it is straightforward to verify that the construction
\begin{equation}\label{eq:H_11}
	E = \frac{1}{2}(H_{11} + \tfrac{1}{\bar\alpha^2}\mathcal{P} + Y_1-Y_1^\top),
\end{equation}
results in $H_{11}=E+E^\top-\tfrac{1}{\bar\alpha^2}\mathcal{P}$ for any $Y_1$.

We then construct a strictly lower-triangular $\mathcal{D}_{11}$ satisfying 
\begin{equation}\label{eq:H22}
	H_{22} = \mathcal{W}=2\Lambda-\mathcal{D}_{11}-\mathcal{D}_{11}^\top
\end{equation} 
by partitioning $H_{22}$ into its diagonal and strictly upper/lower triangular components:
\begin{equation}
	H_{22}=\Phi-L-L^\top
\end{equation} 
where $\Phi$ is a diagonal matrix and $L$ is a strictly lower-triangular matrix, from which we construct the remaining parameters in $\theta_{\mathrm{cvx}}$:
\begin{equation}\label{eq:D11-tri}
	\Lambda=\frac{1}{2} \Phi,\quad \mathcal{D}_{11}=L.
\end{equation}

\paragraph{Constrained parameters, full case}
The construction of a C-REN with full (not acyclic) $D_{11}$ is the same except that we introduce two additional free variables: $g\in \R^q$ and $Y_2 \in \R^{q\times q}$, and then construct a positive diagonal matrix $\Lambda = e^{\mathrm{diag}(g)}$ and
\begin{equation}\label{eq:D11}
	\mathcal{D}_{11} = \Lambda - \frac{1}{2}(H_{22} + Y_2-Y_2^\top),
\end{equation}
which also results \RevF{in} parameters satisfying \eqref{eq:H22}.

\subsection{Direct Parameterizations of Robust RENs}\label{sec:direct_rren}

\Rev{We now provide a direct parameterization of RENs satisfying the robustness condition \eqref{eq:lmi-qsr}. 
The first step is to rearrange \eqref{eq:lmi-qsr} into an equivalent form which will turn out to be useful in the construction since it makes explicit the connection between the R-REN and C-REN conditions:
\begin{subequations}\label{eq:lmi-qsr-2}
	\begin{align}
		&\mathcal{R}:=R+SD_{22}+D_{22}^\top S^\top +D_{22}^\top Q D_{22}\succ 0, \label{eq:lmi-qsr-2-R} \\
		& 
H(\theta_{\mathrm{cvx}})\succ		\begin{bmatrix}
			\mathcal{C}_2^\top \\ \mathcal{D}_{21}^\top \\ \mathcal{B}_2
		\end{bmatrix}\mathcal{R}^{-1}
		\begin{bmatrix}
			\mathcal{C}_2^\top \\ \mathcal{D}_{21}^\top \\ \mathcal{B}_2
		\end{bmatrix}^\top	-
		\begin{bmatrix}
			C_2^\top \\ D_{21}^\top \\ 0 
		\end{bmatrix}Q
		\begin{bmatrix}
			C_2^\top \\ D_{21}^\top \\ 0 
		\end{bmatrix}^\top, \label{eq:lmi-qsr-2-H}
	\end{align}
\end{subequations}
where $H(\theta_{\mathrm{cvx}})$ is the C-REN condition defined in \eqref{eq:lmi-stable}, $\mathcal{C}_2=(D_{22}^\top Q+S )C_2$ and $\mathcal{D}_{21}=(D_{22}^\top Q+S)D_{21}-\mathcal{D}_{12}^\top$.

The first construction we give is for the simplest case without direct-feedthrough, i.e. $D_{22}=0$. }\Rev{However, some practically useful constraints require $D_{22}\ne 0$, e.g., incremental passivity requires $D_{22}+D_{22}^\top \succ 0$. We consider this more general case below.  
\subsubsection{Models with $D_{22} = 0$} for models with no direct feedthrough we have the following direct parameterization.

\paragraph{Free variables} the following matrix variables can be freely parameterized in terms of their elements: $\mathcal{B}_2\in \R^{n\times m}$, $C_2\in \R^{p\times n}$, $\mathcal{D}_{12}\in\R^{q\times m}$, $ D_{21}\in \R^{p\times q}$, $\tilde b \in\R^{(2n+q)}$

\paragraph{Constrained parameters} the construction is similar to the contracting case in Section \ref{sec:direct_caren}.

Since $D_{22}=0$, Condition \eqref{eq:lmi-qsr-2-R} reduces to  $R\succ 0$,  which is independent of model parameters. 
Now Condition~\eqref{eq:lmi-qsr-2-H} can be satisfied if we construct $H$ as 
\begin{equation}\label{eq:H-qsr}
	\begin{split}
		H= &X^\top X +\epsilon I+ \\ &
		\begin{bmatrix}
			\mathcal{C}_2^\top \\ \mathcal{D}_{21}^\top \\ \mathcal{B}_2
		\end{bmatrix}\mathcal{R}^{-1}
		\begin{bmatrix}
			\mathcal{C}_2^\top \\ \mathcal{D}_{21}^\top \\ \mathcal{B}_2
		\end{bmatrix}^\top  -
		\begin{bmatrix}
			C_2^\top \\ D_{21}^\top \\ 0
		\end{bmatrix}Q
		\begin{bmatrix}
			C_2^\top \\ D_{21}^\top \\ 0
		\end{bmatrix}^\top \succ 0,
	\end{split}
\end{equation}
with $X$ a free matrix variable, and then recover the remaining model parameters from $H$ as per Section \ref{sec:direct_caren}. Note that $H\succ 0$,  since $\mathcal R\succ 0$ and $Q\preceq 0$.

\subsubsection{Models with $D_{22} \ne 0$} in this case we need to construct a $D_{22}$ satisfying \eqref{eq:lmi-qsr-2-R}. In what follows it will be useful to have $Q$ invertible but we have only assumed that $Q\preceq 0$. If $Q$ is not negative-definite, we introduce $\mathcal{Q}=Q-\varepsilon I\prec 0$ and note that \eqref{eq:lmi-qsr-2-R} is equivalent to
\begin{equation}\label{eq:qsr-revised}
	R+SD_{22}+D_{22}^\top S^\top +D_{22}^\top \mathcal{Q} D_{22}\succ 0
\end{equation}
for sufficiently small $\varepsilon>0$. If $Q\prec 0$ we simply set $\varepsilon=0$, i.e. $\mathcal{Q}=Q$. 

\Rev{We factor $\mathcal{Q}=-L_Q^\top L_Q$, and we will show \RevF{(see Proposition \ref{prop:D22})} that
$
 R-S\mathcal{Q}^{-1}S^\top \succ 0
$
hence there is an invertible $L_R\in \R^{m\times m}$ such that $L_R^\top L_R=R-S\mathcal{Q}^{-1}S^\top$. 
}

\Rev{The direct parameterization of $D_{22}$ is
\begin{equation}\label{eq:D22}
	D_{22}=-\mathcal{Q}^{-1}S^\top + L_Q^{-1}N L_R,
\end{equation}
where construction of $N$ depends on the input and output dimensions. If  $p\ge m$ we take
\begin{equation}\label{eq:M}
	\begin{split}
	    M&=X_3^\top X_3+Y_3-Y_3^\top + Z_3^\top Z_3+\epsilon I, \\
        N&=\begin{bmatrix}
        (I-M)(I+M)^{-1} \\
        -2Z_3(I+M)^{-1}
	    \end{bmatrix}.
 \end{split}
\end{equation}
with $X_3, Y_3\in \R^{m\times m}$ and $Z_3\in \R^{(p-m)\times m}$ as  free variables. Note that $M+M^\top\succ 0$ so $\RW{I+M}$ is invertible.

If $p<m$, $M$ is the same but we take
\begin{equation}\label{eq:M2}
    N=\begin{bmatrix}
	        (I+M)^{-1}(I-M) &
	        -2(I+M)^{-1}Z_3^\top
	    \end{bmatrix}
\end{equation}
with $X_3, Y_3\in \R^{p\times p}$ and $Z_3\in \R^{(m-p)\times p}$ as free variables.
}

\begin{proposition}\label{prop:D22}
	The construction of $D_{22}$ \Rev{in} \eqref{eq:D22}, \eqref{eq:M} or \eqref{eq:M2} is well-defined and satisfies Condition \eqref{eq:qsr-revised}.
\end{proposition}
The proof is in Appendix~\ref{sec:proof-prop-D22}.

\paragraph{Special Cases}
the following are direct parameterizations of $D_{22}$ for some commonly-used robustness conditions: 
\begin{itemize}
	\item Incrementally $\ell_2$ stable RENs with Lipschitz bound of $\gamma$ (i.e., $Q=-\frac{1}{\gamma}I, R=\gamma I, S=0$)\Rev{: We have $D_{22}$ given in \eqref{eq:D22} with $L_Q=I$ and $L_R=\gamma I$.}
	
	\item Incrementally strictly output passive RENs (i.e., $Q=-2\rho I,R=0,S=I$): We have $ D_{22}=\frac{1}{\rho}(I+M)^{-1}$.
	\item Incrementally input passive RENs (i.e., $Q=0,R=-2\nu I,S=I$): In this case, Condition~\eqref{eq:lmi-qsr-2-R} becomes an LMI of the form $ D_{22}+D_{22}^\top -2\nu I\succ 0$, which yields a simple parameterization with $ D_{22}=\nu I +M$.
\end{itemize}}

\subsection{Random Sampling of Nonlinear Systems and Echo State Networks}\label{sec:echostate}

One benefit of the direct parameterizations of RENs is that it is straightforward to randomly sample systems with the desired behavioural properties. Since contracting and robust RENs are constructed as the image of $\mathbb R^N$ under a smooth mapping (Sections \ref{sec:direct_caren} and \ref{sec:direct_rren}), one can sample random vectors in $\R^N$ and map them to random stable/robust nonlinear dynamical systems.

An ``echo state network'' is a model in which the state-space dynamics are randomly sampled but thereafter fixed, and with a learnable output map (see ,e.g., \cite{buehnerTighterBoundEcho2006, yildizRevisitingEchoState2012}):
\begin{align}
    x_{t+1} &= f(x_t, u_t)\label{eq:esn_x}\\
    y_{t+1} &= g(x_t,u_t ,\theta)
\end{align}
where $f$ is fixed and $g$ is affinely parameterized by $\theta$, i.e.
\[
g(x_t,u_t ,\theta)=g_0(x_t,u_t)+\sum_i \theta_i g^i(x_t,u_t ).
\]
Then, system identification with a simulation-error criteria can be solved as a basic least squares problem. This approach is reminiscent of system identification via a basis of stable linear responses (see, e.g., \cite{wahlbergApproximationStableLinear1996}). 

For this approach to work over long horizons, it is essential that the random dynamics are stable. In \cite{buehnerTighterBoundEcho2006, yildizRevisitingEchoState2012} and references therein, contraction of \eqref{eq:esn_x} is referred to as the ``echo state property'', and simple parameterizations are given for which contraction is guaranteed. The direct parameterizations of REN can be used to randomly sample from a rich class of contracting models, by sampling $X, Y_1, Y_2, \mathcal{B}_2, \mathcal{D}_{12}$ to construct the state-space dynamics and equilibrium network. Such a model can be used e.g. for system identification by simulating its response to inputs to generate data $\tilde u_t, \tilde x_t, \tilde w_t$, and then the output mapping
\[
y_t = C_2\tilde x_t+D_{21}\tilde w_t+D_{22}\tilde u_t+b_y
\]
can be fit to $\tilde y_t$, minimizing \eqref{eq:sim_error} via least-squares to obtain the parameters $C_2, D_{21}, D_{22}, b_y$. We will also see in Section \ref{sec:youla} how this approach can be applied in data-driven feedback control design.

\section{Expressivity of REN Model Class}\label{sec:express}
The set of RENs contain many widely-used model structures as special cases, some of which we briefly describe here.

\paragraph{Deep, Residual, and Equilibrium Networks} as a special case with $A, C_1, C_2, B_1, B_2$ all zero, RENs include (static) equilibrium networks, which as discussed in Section \ref{sec:feedforward} and \cite{ghaoui2019implicit, winston2020monotone, revay2020lipschitz} include standard deep neural networks (multi-layer perceptrons), residual networks, and others.

\paragraph{Previously proposed stable RNNs} if we set $D_{11} = 0$, then the nonlinearity is not an equilibrium network but a single-hidden-layer neural network, and our model set $\Theta_C$ reduces to the model set proposed in \cite{revay2020convex}. Therefore, the REN model class also includes all other models that were proven to be in that model set in \cite[Theorem 5]{revay2020convex}, including prior sets of contracting RNNs including the ciRNN \cite{revay2020contracting} and s-RNN\cite{miller2018stable}.

\paragraph{Stable linear systems} setting $B_1, C_1, D_{11}, D_{12}, D_{21}$ and $b$ to zero, RENs include all stable finite-dimensional linear time-invariant (LTI) systems (see \cite[Theorem 4]{revay2020convex}).

\paragraph{Previously proposed stable echo state networks} the stability condition for the ciRNN is the same as that proposed for echo state networks in \cite{buehnerTighterBoundEcho2006,yildizRevisitingEchoState2012}, hence by randomly sampling RENs as in Section \ref{sec:echostate} we sample from a strictly larger set of echo state networks than previously known.

\paragraph{Nonlinear finite impulse response (NFIR) Models} an NFIR model a nonlinear mapping of a fixed history of inputs:
\[
y_t = f(u_t, u_{t-1}, ... u_{t-h}),
\]
for some fixed $h$. Setting 
\begin{gather}
A = 
\begin{bmatrix}
0 &   & & \\
I & 0 & & \\
& I &\sddots& \\
&   &\sddots& 
\end{bmatrix}, \quad B_2 = \begin{bmatrix}
I \\ 0 \\ 0 \\ \svdots
\end{bmatrix}, \quad B_1 = 0.
\end{gather}
The output $y$ is then a nonlinear function (an equilibrium network) of such truncated history of inputs. 

\paragraph{Block structured models} these are constructed from series interconnections of LTI systems and static nonlinearities \cite{schoukens2017block,giri2010block}, and are included within the REN model set. For example:
\begin{enumerate}
	\item \textit{Wiener systems} consist of an LTI block followed by a static non-linearity. This structure is replicated in \eqref{eq:G}, \eqref{eq:sigma} when $B_1 = 0$ and $C_2 = 0$. In this case the linear dynamical system evolves independently of the non-linearities and feeds into a equilibrium network.
	\item \textit{Hammerstein systems} consist of a static non-linearity connected to an LTI system. This is represented in the REN when $B_2 = 0$ and $C_1 = 0$. In this case the input passes through a static equilibrium network and into an LTI system.
\end{enumerate}
More generally, arbitrary series and parallel interconnections of LTI systems and static nonlinearities can also be constructed.

\paragraph{Universal approximation properties} it is well known even single-hidden-layer neural networks have universal approximation properties, i.e. as the number of neurons goes to infinity they can approximate any continuous function over a bounded domain with arbitrary accuracy. RENs immediately inherit this property for universal approximation of static maps, NFIR models, and other block-structured models.

Furthermore, it was shown in \cite{wang2022learning} that as the number of states and activation functions grows, the REN structure is a universal approximator of fading-memory nonlinear systems as defined in \cite{boyd1985fading}, as well as all nonlinear dynamical systems that are contracting and have finite Lipschitz bounds.

\section{Use Case: Stable and Robust Nonlinear System Identification}
\label{sec:ID}
In this section we demonstrate the proposed models on the F16 ground vibration \cite{schoukens2017f16} and Wiener Hammerstein with process noise  \cite{schoukens2017wiener} system identification benchmarks. We will compare the acyclic C-REN and Lipschitz-bounded R-REN with prescribed Lipschitz bound of $\RW{\gamma} $ with the widely-used long short-term memory (LSTM) \cite{Hochreiter:1997} and standard RNN models with a similar number of parameters. We will also compare to the Robust RNN proposed in \cite{revay2020convex} using the code from \url{github.com/imanchester/RobustRNN}.

We fit models by minimizing simulation error:
\begin{equation}\label{eq:sim_err}
\mathcal{L}_{se}(\tilde z,\theta) = ||\tilde{y} - \mathfrak{R}_a(\tilde{u})||^2_T
\end{equation}
using minibatch gradient descent with the Adam optimizer \cite{Kingma:2017}. Model performance is measured by normalized root mean square error on the test sets, calculated as:
\begin{equation}\label{eq:NRMSE}
\text{NRMSE} = \frac{||\tilde{y} - \mathfrak{R}_a(\tilde{u})||_T}{||\tilde{y}||_T}.
\end{equation}
Model robustness is measured in terms of the maximum observed sensitivity:
\begin{equation} \label{eq:lipschitz_lower}
\underline{\gamma}  = \underset{u, v, a}{\max} \frac{||\mathfrak{R}_a(u)-\mathfrak{R}_a(v)||_T}{||u - v||_T}.
\end{equation}
We find a local solution to \eqref{eq:lipschitz_lower} using gradient ascent with the Adam optimizer. Consequently $\underline{\gamma}$ is a lower bound on the true Lipschitz constant of the sequence-to-sequence map.  

\subsection{Benchmark Datasets and Training Details}

\subsubsection{F16 System Identification Benchmark}
The F16 ground vibration benchmark  dataset \cite{schoukens2017f16} consists of accelerations measured by three accelerometers, induced in the structure of an F16 fighter jet by a wing mounted shaker. We use the multi-sine excitation dataset with full frequency grid. This dataset consists of 7 multi-sine experiments with 73,728 samples and varying amplitude. We use datasets 1, 3, 5 and 7 for training and datasets 2, 4 and 6 for testing. 

All models in our comparison  have approximately 118,000 parameters: the RNN has 340 neurons, the LSTM has 170 neurons and the RENs have width $n=75$ and $q = 150$.
Models were trained for $70$ epochs with a sequence length of $1024$. The learning rate was initalized at $10^{-3}$ and was reduced by a factor of $10$ every $20$ Epochs. 

\subsubsection{Wiener-Hammerstein With Process Noise Benchmark}

The Wiener Hammerstein with process noise benchmark dataset \cite{schoukens2017wiener} involves the estimation of the output voltage from two input voltage measurements from a Wiener-Hammerstein system with large process noise. We have used the multi-sine fade-out dataset consisting of two realisations of a multi-sine input signal with 8192 samples each. The test set consists of two experiments, a random phase multi-sine and a sine sweep, 
conducted without the added process noise. 

All models in our comparison  have approximately 42,000 parameters: the RNN has 200 neurons, the LSTM has 100 neurons and the RENs have $n=40$ and $q=100$. Models were trained for 60 epochs with a sequence length of $512$. The initial learning rate was $10^{-3}$ and was reduced to $10^{-4}$ after $40$ epochs.

\subsection{Results and Discussion}\label{sec:sysid results}
\begin{figure}
	\centering
    \includegraphics[width=0.8\linewidth, trim={0cm, 0cm, 0cm, 0cm},clip]{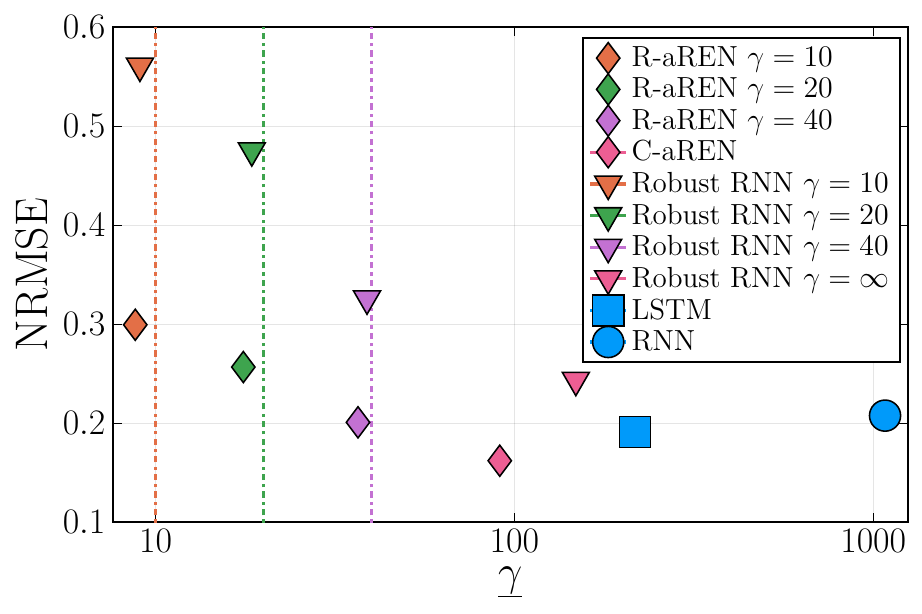}
	\caption{\label{fig:f16_results} Nominal performance versus robustness for models trained on F16 ground vibration benchmark dataset. The dashed vertical lines are the guaranteed upper bounds on $\gamma$ corresponding to the models with matching color.}
\end{figure}

\begin{figure}
	\centering
	\includegraphics[width=0.8\linewidth, trim={0cm, 0cm, 0cm, 0cm},clip]{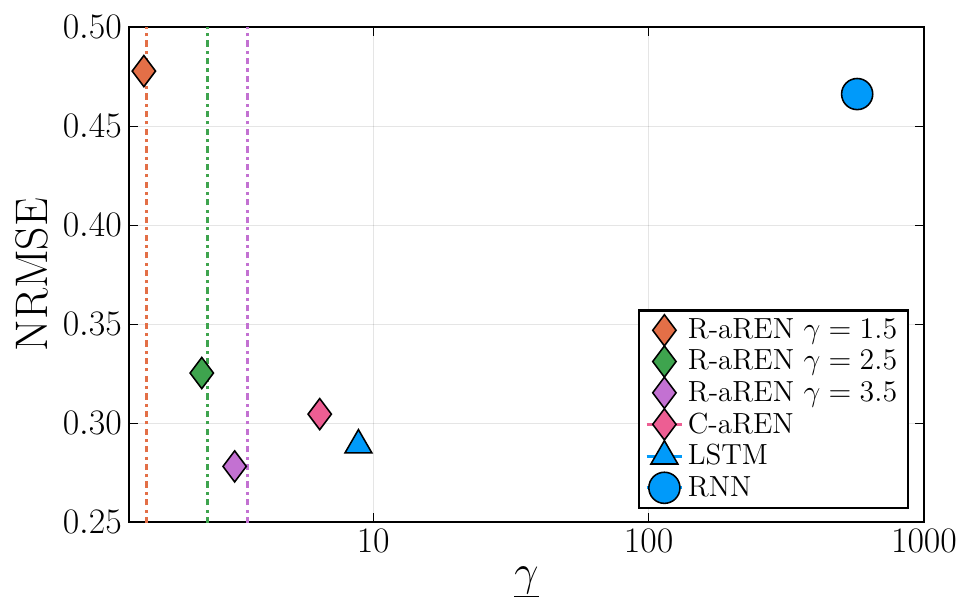}
	\caption{\label{fig:wh_results} Nominal performance versus robustness for models trained on Wiener-Hammerstein with process noise benchmark dataset. The dashed vertical lines are the guaranteed upper bounds on $\gamma$ corresponding to the models with matching color.}
\end{figure}

In Figs.~\ref{fig:f16_results} and \ref{fig:wh_results} we have plotted the test-set NRMSE \eqref{eq:NRMSE} versus the observed sensitivity \eqref{eq:lipschitz_lower} for each of the models trained on the F16 and Wiener-Hammerstein Benchmarks, respectively. The dashed vertical lines show the guaranteed Lipschitz bounds for the REN and Robust RNN models. 

We observe that the REN offers the best trade-off between nominal performance and robustness, with the REN slightly outperforming the LSTM in terms of nominal test error for large $\gamma$. By tuning $\gamma$, nominal test performance can be traded-off for robustness, signified by the consistent trend moving diagonally up and left with decreasing $\gamma$. In all cases, we found that the REN was significantly more robust than the RNN, typically having about $10\%$ of the sensitivity for the F16 benchmark and $1\%$ on the Wiener-Hammerstein benchmark. Also note that for small $\gamma$, the observed lower bound on the Lipschitz constant is very close to the guaranteed upper bound, showing that the real Lipschitz constant of the models is close to the upper bound. 

Compared to the robust RNN proposed in \cite{revay2020convex}, the REN has similar bounds on the incremental $\ell_2$ gain, however the added flexibility from the term $D_{11}$ significantly improves the nominal model performance for a given gain bound. Additionally, while both the C-REN and Robust RNN $\gamma\RW{=}\infty$ are contracting models, we note that the C-REN is significantly more expressive with a NRMSE of 0.16  versus 0.24.

\begin{figure}
	\centering
	\includegraphics[width=0.8\linewidth, trim={0cm, 0cm, 0cm, 0cm},clip]{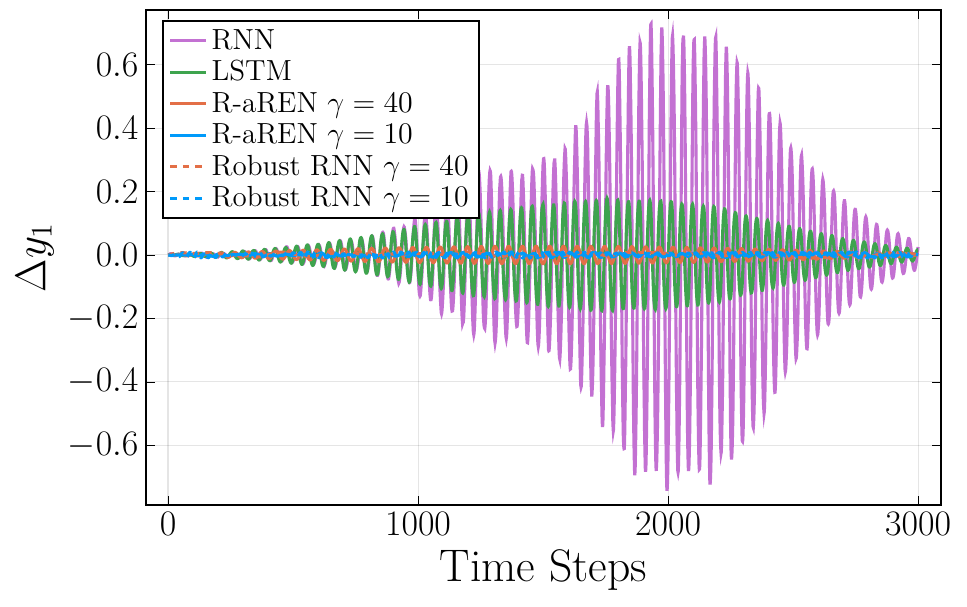}
	\caption{\label{fig:f16_adversarial_perturbation} Change in output of models subject to an adversarial perturbation with $||\Delta u|| < 0.05.$ The incremental gains from $\Delta u$ to $\Delta y$ are 980, 290, 37, 8.6, \Rev{38.9 and 9.1}, respectively.}
\end{figure}

\begin{figure}
	\centering
	\includegraphics[width=0.8\linewidth, trim={0cm, 0cm, 0cm, 0cm},clip]{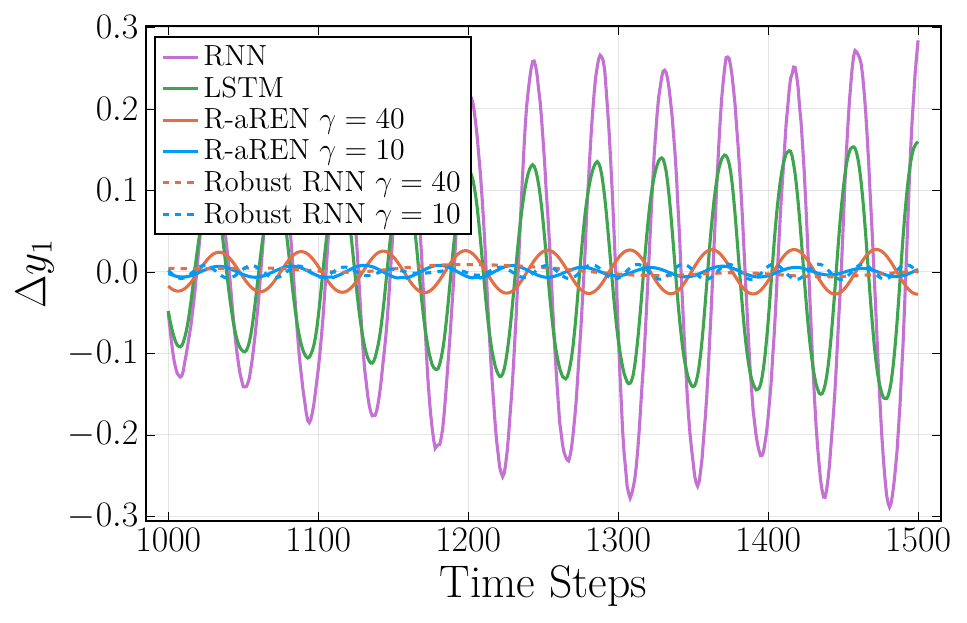}
	\caption{\label{fig:f16_adversarial_perturbation_zoomed} Zoomed in version of Fig. \ref{fig:f16_adversarial_perturbation}. }
\end{figure}

It is well known that many neural networks are very sensitive to adversarial perturbations. This is shown, for instance, in Fig.~\ref{fig:f16_adversarial_perturbation} and \ref{fig:f16_adversarial_perturbation_zoomed}, where we have plotted the change in output for a small adversarial perturbation $||\Delta u||<0.05$, for a selection of models trained on the F16 benchmark dataset. Here, we can see that both the RNN and LSTM are very sensitive to the input perturbation. The R-REN \Rev{and R-RNN} on the hand, have guaranteed bounds on the effect of the perturbation and are significantly more robust.

We have also trained \Rev{cyclic RENs (i.e. $D_{11}$ is a full matrix)} for the F16 Benchmark dataset. The resulting nominal performance and sensitivities for the \Rev{acyclic and cyclic} RENs are shown in Table~\ref{tab:REN}. We do not observe a significant difference in performance between the cyclic and acyclic model classes.

\begin{figure}
	\centering
	\includegraphics[width=0.8\linewidth, trim={0cm, 0cm, 0cm, 0cm},clip]{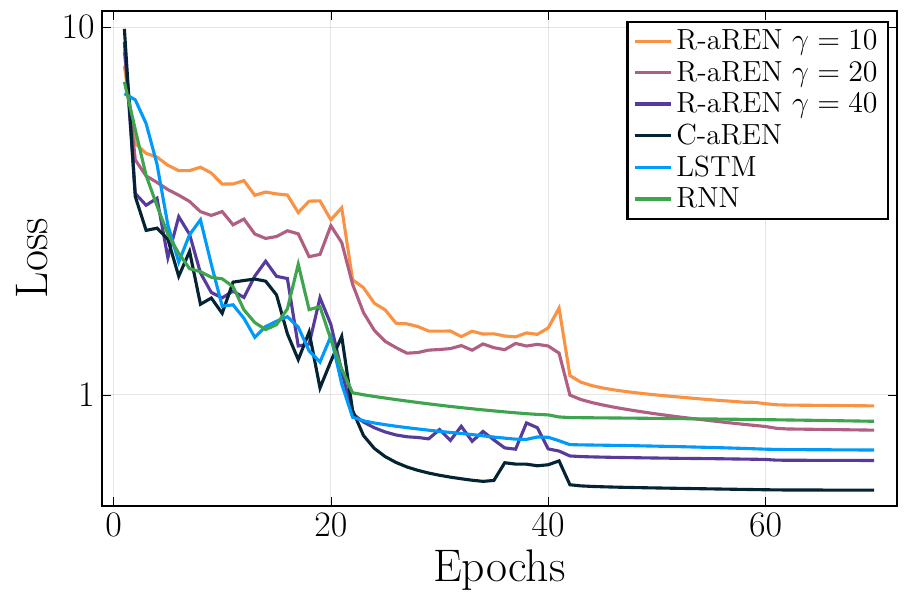}
	\caption{\label{fig:f16_training} Traing loss versus epochs for models trained on F16 ground vibration benchmark dataset.}
\end{figure}

\begin{table}[]
	\centering
	\caption{\label{tab:REN} Nominal performance (NRMSE) and  upper and lower bounds on Lipschitz constant for \Rev{acyclic and cyclic} RENs \RW{on F16 benchmark dataset}.}
	\renewcommand{\arraystretch}{1.6}
	\begin{tabular}{|c|c||c|c|c|c|c |c|}\hline
 & $\RW{\gamma}$& 10 & 20 & 40 & 60 & 100 & $\infty$\\ \hline \hline
		\multirow{2}{*}{\Rev{acyclic}}		& $\underline{\gamma}$  & 8.8 & 17.5 & 36.7 & 44.9 & 60.56 & 91.0\\ \cline{2-8}
		& NRMSE ($\%$) & 30.0 & 25.7 & 20.1 & 18.5 & 17.2 & 16.2\\ \hline
		\multirow{2}{*}{\Rev{cyclic}} & $\underline{\gamma}$ & 9.1& 17.1 & 36.0& 44.6 & 57.9 & 85.26  \\ \cline{2-8}
		& NRMSE ($\%$) &  30.3 & 26.8 & 21.8 & 19.9 & 19.3 & 16.8 \\ \hline
	\end{tabular}
\end{table}
Finally, we have plotted the training loss \eqref{eq:sim_err} versus the number of epochs in Fig. \ref{fig:f16_training} for some of the models on the F16 dataset. Compared to the LSTM, the REN takes a similar number of steps and achieves a slightly lower training loss. 

\section{Use Case: Learning Nonlinear Observers}
\label{sec:obs}
Estimation of system states from incomplete and/or noisy measurements is an important problem in many practical applications. For linear systems with Gaussian noise, a simple and optimal solution exists in the form of the Kalman filter, \RevF{but for nonlinear systems even finding a stable estimator (a.k.a. observer) is non-trivial and many approaches have been investigated}, e.g. \cite{astolfi2007nonlinear, khalil2017high, bernard2019observer}. Observer design was one of the original motivations for contraction analysis \cite{Lohmiller:1998}, and in this section, we show how a flexible set of contracting models can be used to learn stable state observers via \textit{snapshots} of a nonlinear system model.

The aim is to estimate the  state of a nonlinear system of the form 
\begin{equation}
  x_{t+1} = f_m(x_t,u_t,w_t), \quad y_t=g_m(x_t,u_t,w_t)
\end{equation}
where $x_t\in \mathbb{X}$ is an internal state to be estimated, $y_t$ is an available measurement, $u_t\in \mathbb{U}$ is a known (e.g. control) input, and $w_t\in\mathbb{W}$ comprises unknown disturbances and sensor noise.

A standard structure, pioneered by Luenberger, is an observer of the form
\begin{equation}\label{eq:luenberger}
  \hat x_{t+1} = f_m(\hat x_t,u_t,0)+l(\hat x_t, u_t, y_t)
\end{equation}
i.e. a combination of a model prediction $f_m$ and a measurement correction function $l$. A common special case is $l(\hat x_t, u_t, y_t)=L(\hat x)(y_t-g_m(\hat x_t,u_t,0))$ for some gain $L(\hat x)$.

In many practical cases the best available model $f_m, g_m$ is highly complex, e.g. based on finite element methods or algorithmic mechanics \cite{featherstone2014rigid}. This poses two major challenges to the standard paradigm:
\begin{enumerate}
  \item How to design the function $l$ such that the observer \eqref{eq:luenberger} is stable (preferably globally) and exhibits good noise/disturbance rejection.
  \item The model itself may be so complex that evaluating $f_m(\hat x_t,u_t,0)$ in real-time is infeasible, e.g. for stiff systems where short sample times are required.
\end{enumerate}
Our parameterization of contracting models enables an alternative paradigm, first suggested for the restricted case of polynomial models in \cite{manchester2018contracting, yi2020reduced}.

\begin{proposition}\label{prop:2}
	If we construct an observer of the form
\begin{equation}\label{eq:contracting_observer}
  \hat x_{t+1} = f_o(\hat x_t,u_t,y_t)
\end{equation}
such that the following two conditions hold:
\begin{enumerate}
  \item The system \eqref{eq:contracting_observer} is contracting with rate $\alpha \in (0,1)$ for some constant metric $P\succ 0$.
  \item The following ``correctness'' condition holds:
  \begin{equation}\label{eq:correctness}
  f_m(x,u,0)=f_o(x,u,g_m(x,u,0)),\; \forall (x,u)\in \mathbb{X}\times\mathbb{U}.
\end{equation}
\end{enumerate}
Then when $w=0$ we have $\hat x_t\to x_t$ as $t\to \infty$. Suppose instead Condition 2) does not hold but that the observer \eqref{eq:contracting_observer} satisfies Conditions 1) and 
\begin{enumerate}
  \item[3)] The following error bound holds $\forall (x,u,w)\in \mathbb{X}\times\mathbb{U}\times\mathbb{W}$:
  \begin{equation}\label{eq:obs_fit_err}
    |f_o(x,u,g_m(x,u,w))-f_m(x,u,w)|\le \rho. 
  \end{equation} 
\end{enumerate}  
Then the estimation error satisfies, with exponential convergence:
\begin{equation}\label{eq:estimation-bound}
\lim\sup_{t\rightarrow \infty} |\hat{x}_t-x_t|\leq \frac{\rho}{1-\alpha}\sqrt{\frac{\overline{\sigma}}{\underline{\sigma}}},
\end{equation}
where $\overline{\sigma}$ and $\underline{\sigma}$ denote the maximum and minimum singular values of the contraction metric $P$, respectively.
\end{proposition} 

\RevF{\begin{remark} Note that the error term \eqref{eq:obs_fit_err} may result from bounded disturbances $w_t$, modelling errors, or interpolation errors arising from fitting the correctness condition to finite data (see Sec \ref{sec:obs_example}), or some combination of such factors.  
\end{remark}}

The reasoning for nominal convergence of the observer is simple: \eqref{eq:correctness} implies that if $\hat x_0=x_0$ then $\hat x_t=x_t$ for all $t\ge 0$, i.e. the true state is a particular solution of the observer. But contraction implies that all solutions of the observer converge to each other. Hence all solutions of the observer converge to the true state. The proof of the estimation error bound can be found in Appendix~\ref{sec:proof-prop-2}.

Motivated by Proposition \eqref{prop:2} we pose the observer design problem as a supervised learning problem over our class of contracting models.

\begin{enumerate}
  \item Construct the dataset: sample a set of points $\tilde z = \{(x^i, u^i), i=1, 2, ..., N\}$ where $ (x^i,u^i)\in \mathbb{X}\times \mathbb{U}$, and for each compute $g_m^i = g_m(x^i,u^i,0)$ and $f_m^i = f_m(x^i,u^i,0)$.
  \item Learn a contracting system $f_o$ minimizing the loss \begin{equation}\label{eq:obs_loss}
  \mathcal{L}_o(\tilde z,\theta) = \sum_{i=1}^N \left|f^i_m - f_o(x^i,u^i, g^i_m)\right|^2.
\end{equation}
\end{enumerate}
 
\begin{remark}
	An observer of the traditional form \eqref{eq:luenberger} with $l(\hat x_t, u_t, y_t)=L(\hat x)(y_t-g_m(\hat x_t,u_t,0))$ will always satisfy the correctness condition, but designing $L(\hat x)$ to achieve global convergence may be difficult. In contrast, an observer design using the proposed procedure will always achieve global convergence, but may not achieve correctness exactly. 
\end{remark}

\subsection{Example: Reaction-Diffusion PDE}\label{sec:obs_example}
We illustrate this approach by designing an observer for the following semi-linear reaction-diffusion partial differential equation:
\begin{gather}
  \frac{\partial \xi(z,t)}{\partial t}= \frac{\partial^2 \xi(z,t)}{\partial z^2}+R(\xi,z,t), \label{eq:reaction_diffusion}\\
  \xi(z, 0) = 1, \quad \xi(1, t) = \xi(0, t) = b(t) \\
  y = g(\xi, z, t)
\end{gather}
where the state $\xi(z,t)$ is a function of both the spatial coordinate $z\in[0,1]$  and time $t\in\R_+$. 
Models of the form \eqref{eq:reaction_diffusion} model processes such as combustion \cite{gilding2012travelling}, bioreactors \cite{meurer2013extended} or neural spiking dynamics \cite{gilding2012travelling}. The observer design problem for such systems has been considered using complex back-stepping methods that guarantee only local stability \cite{meurer2013extended}. 

We consider the case where the local reaction dynamics have the following form, which appears in models of combustion processes \cite{gilding2012travelling}: 
\[R(\xi, z, t) = \frac{1}{2}\xi(1-\xi)(\xi-\tfrac{1}{2}).\]
 We consider the boundary condition $b(t)$ as a known input and assume that there is a single measurement taken from the center of the spatial domain so $y(t) = \xi(0.5, t)$.

We discretize $z$ into $N$ intervals with points $z^0,...,z^{N}$ where $z^i = i\Delta z$. The state at spatial coordinate $z^i$ and time $t$ is then described by $\bar{\xi}_t = (\xi^0_t, \xi^1_t, ..., \xi^N_t)$ where $\xi^i_t = \xi(z^i, t)$.
The dynamics over a time period $\Delta t$ can then be approximated using the following finite differences:
\[
    \frac{\partial \xi(z,t)}{\partial t} \approx   \frac{\xi^i_{t+\Delta t} - \xi^i_t}{\Delta t}, \quad
	 \frac{\partial^2 \xi(z,t)}{\partial z^2} \approx \frac{\xi^{i+1}_t + \xi^{i-1}_t - 2\xi^i_t}{\Delta z^2}.
\]
Substituting them into \eqref{eq:reaction_diffusion} and rearranging for $\bar{\xi}_{t+\Delta t}$ leads to an $N+1$ dimensional state-space model of the form:
\begin{equation}\label{eq:reaction_diffusion_discrete_dynamics}
    \bar{\xi}_{t+\Delta t} = a_{rd}(\bar{\xi}_t, b_t),\quad y_t = c_{rd}(\bar{\xi}_t).
\end{equation}
We generate training data by simulating the system  \eqref{eq:reaction_diffusion_discrete_dynamics} with $N=50$ for $10^5$ time steps with the stochastic input $b_{t+1} = b_t + 0.05 \omega_t$ where $\omega_t \sim \mathcal{N}[0,1]$. We denote this training data by  $\tilde{z} = (\tilde{\xi}_t, \tilde{y}_t, \tilde{b}_t)$ for $t=0,\ldots,10^5\Delta t$.

To train an observer for this system, we construct a C-REN with $n=51$ and $q = 200$. We optimize the one step ahead prediction error:
$$
\mathcal{L}(\tilde{z}, \theta) = \frac{1}{T}\sum_{t=0}^{T-1} |\Rev{a_{rd}}(\tilde{\xi}_t, \tilde{b}_t) - f_o(\tilde{\xi}_t, \tilde{b}_t, \tilde{y}_t)|^2,
$$
using SGD with the Adam optimizer \cite{Kingma:2017}. Here, $f_o(\xi, b, y)$ is a C-REN described by \eqref{eq:G}, \eqref{eq:sigma} using direct parametrization discussed in Section \ref{sec:direct_caren}. Note that we have taken the output mapping in \eqref{eq:G} to be $[C_2, D_{21}, D_{22}] = [I, 0, 0]$.

\begin{figure}[]
	\centering
\includegraphics[width=0.8\columnwidth]{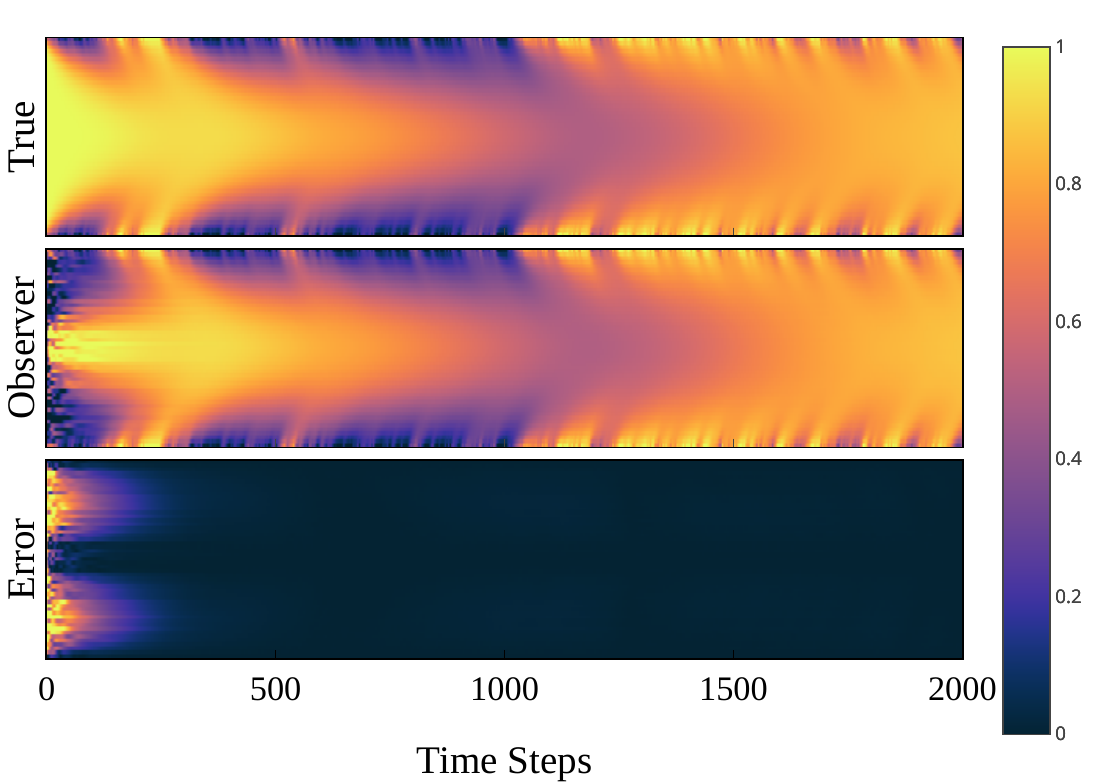}
  \caption{\label{fig:pde_observer_image}Simulation of a semi-linear reaction diffusion equation and the observer's state estimate, with a measurement in the centre of the spatial domain. The $y$-axis corresponds to the spatial dimension and the $x$-axis corresponds to the time dimension.}
\end{figure}

We have plotted results of the PDE simulation and the observer state estimates in Fig. \ref{fig:pde_observer_image}. The simulation starts with an initial state of $\xi(z, 0) = 1$  and the observer has an initial state estimate of $\bar{\xi}_0 = 0$. The error between the state estimate and the PDE simulation's state quickly decays to zero and the observer state continues to track the PDE's state.

\begin{figure}[hbt]
\centering
	\begin{subfigure}{\linewidth}
		\centering
  \includegraphics[width=0.8\columnwidth]{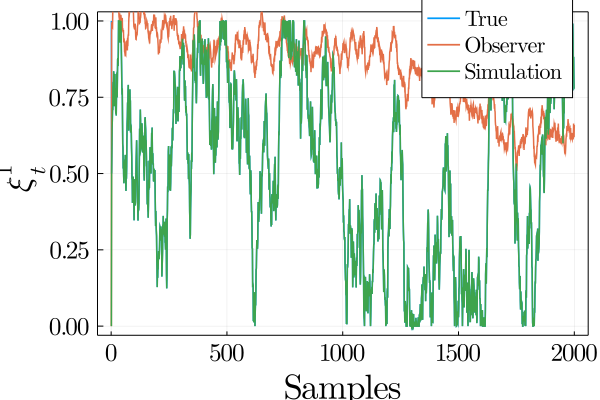}
  \caption{True and estimated states for $\xi^{1}_t$, located at PDE boundary.}
  \end{subfigure}
	\begin{subfigure}{\linewidth}
		\centering
  \includegraphics[width=0.8\columnwidth]{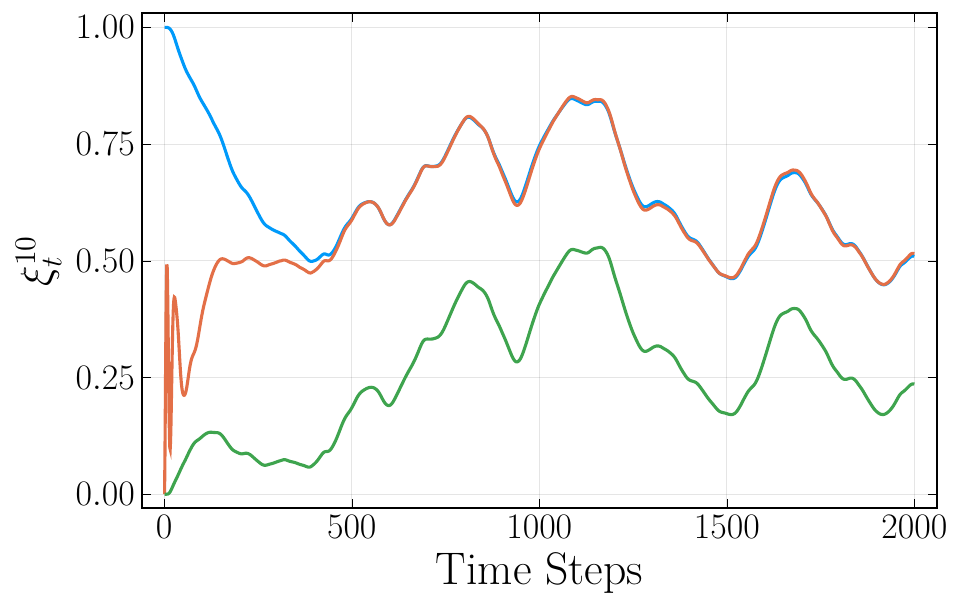}
	\caption{True and estimated states for $\xi^{10}_t$.}
    \end{subfigure}
\caption{\label{fig:pde_observer_states}True state and state estimates from the designed observer and a free run simulation of the PDE.}
\end{figure}


We have also provided a comparison to a free run simulation of the PDE with initial condition $\xi(z, 0) = 0$ in Fig.~\ref{fig:pde_observer_states}. Here we can see that simulated trajectories with different initial conditions do not converge. This suggests that the system is not contracting and the state cannot be estimated by simply running a parallel simulation. The state estimates of the observer, however, quickly converge on the true state.

\section{Use Case: Data-Driven Feedback Control Design}
\label{sec:youla}

\newcommand{\Am}{\mathbb A}
\newcommand{\Bm}{\mathbb B}
\newcommand{\Hm}{\mathbb H}
\newcommand{\Cm}{\mathbb C}
\newcommand{\Dm}{\mathbb D}
\newcommand{\Km}{\mathbb K}
\newcommand{\Lm}{\mathbb L}

\newcommand{\Pb}{P}
\newcommand{\Qb}{\mathcal{Q}}

In this section we show how a rich class of contracting nonlinear models can be useful for nonlinear feedback design for \textit{linear} dynamical systems with stability guarantees. Even if the dynamics are linear, the presence of constraints, uncertain parameters, non-quadratic costs, and non-Gaussian disturbances can mean that non-linear policies  are superior to linear policies. Indeed, in the presence of constraints, model predictive control (a nonlinear policy) is a common approach.

The basic idea we illustrate in this section is to build on a standard method for linear feedback optimization: the Youla-Kucera parameterization, a.k.a Q-augmentation \cite{youlaModernWienerHopfDesign1976, boyd1991linear, zhou1996robust, hespanhaLinearSystemsTheory2018}. For a discrete-time linear system model
\begin{align}
  x_{t+1}&=\Am x_t+\Bm_1 w_t+\Bm_2 u_t, \\
   \zeta_t&= \Cm_{1} x_t+\Dm_{11}w_t+\Dm_{12} u_t. \\
  y_t& = \Cm_2 x_t+\Dm_{21}w_t.
\end{align}
with $x$ the state, $u$ the controlled input, $w$  external inputs (reference, disturbance, measurement noise), $y$ a measured output, and $\zeta$ comprises the ``performance'' outputs to kept small (e.g. tracking error, control signal). We assume the system is detectable and stabilizable, i.e. there exist $\Lm$ and $\Km$ such that $\Am-\Lm\Cm$ and $\Am-\Bm\Km$ are Schur stable. Note that if $\Am$ is stable we can take $\Lm=0, \Km=0$. Consider a feedback controller of the form:
\begin{align}
  \hat x_{t+1} &= \Am \hat x_t+\Bm_2 u_t+\Lm\tilde y\\
  \tilde y_t&=y_t-\Cm_2\hat x_t\\
  u_t&=-\Km\hat x_t+\tilde u_t
\end{align}
i.e. a standard output-feedback structure with $v_t$ an additional control augmentation. The closed-loop input-output dynamics can be written as the transfer matrix 

\begin{equation}
  \begin{bmatrix}
  	\zeta\\\tilde y
  \end{bmatrix}
  =
    \begin{bmatrix}
        \mathcal{T}_0 & \mathcal{T}_1 \\
        \mathcal{T}_2 & 0
    \end{bmatrix}
  \begin{bmatrix}
  	w\\ \tilde u
  \end{bmatrix}
\end{equation}
where we have used the fact that $\tilde u$ maps to $x$ and $\hat x$ equally, hence the mapping from $\tilde u$ to $\tilde y$ is zero.

It is well-known that the set of all stabilizing linear feedback controllers can be parameterised by stable linear systems $\Qb:\tilde y\mapsto \tilde u$, and moreover this convexifies the closed-loop dynamics. A standard approach (e.g. \cite{boyd1991linear, hespanhaLinearSystemsTheory2018}) is to construct an affine parameterization for $\Qb$ via a finite-dimensional truncation of a complete basis of stable linear systems, and optimize to meet various criteria on frequency response, impulse response, and response to application-dependent test inputs. However, if the control augmentation $\tilde u$ is instead generated by a contracting nonlinear system $\tilde u=\Qb(\tilde y)$, then the closed-loop dynamics $w\mapsto \zeta$ are nonlinear but contracting and have the representation
\begin{equation}
    \zeta = \mathcal{T}_0w+\mathcal{T}_1\Qb (\mathcal{T}_2 w)
\end{equation}
This presents opportunities for learning stabilizing controllers via parameterizations of stable nonlinear models.

\subsection{Echo State Network and Convex Optimization}

Here we describe a particular setting in which the data-driven optimization of nonlinear policies can be posed as a \textit{convex} problem. Suppose we wish to design a controller solving (at least approximately) a problem of the form:
\begin{equation}
    \min_\theta J(\zeta)\quad  \textrm{s.t.}\quad c(\zeta)\le 0
\end{equation}
where $\zeta$ is the response of the performance outputs to a \textit{particular class} of disturbances $w$,  $J$ is a convex objective function, and $c$ is a set of convex constraints, e.g. state and control signal bounds.

If we take $\Qb$ as an echo state network, c.f. Section \ref{sec:echostate}:
\[
    q_{t+1} = f_q(q_t,\tilde y_t), \quad \tilde u_{t} = g_q(q_t,\tilde y_t,\theta)
\]
where $f_q$ is fixed and $g_q$ is linearly parameterized by $\theta$, i.e.
\[
g_q(q_t,\tilde y_t,\theta)=\sum_i \theta_i g_q^i(q_t,\tilde y_t).
\]
Then $\Qb$ has the representation
\[
\Qb(\tilde y) = \sum_i \theta_i \Qb^i(\tilde y)
\]
where $\Qb^i$ is a state-space model with dynamics $f_q$ and output $g_q^i$. Then, we can perform data-driven controller optimization in the following way:
\begin{enumerate}
    \item Construct (e.g. via random sampling, experiment) a finite set of test signals $w^j$.
    \item Compute $\tilde y_t^j = \mathcal{T}_2 w^j$ for each $j$.
    \item For each $j$, compute the response to $\tilde y^j$: \[q_{t+1} = f_q(q_t,\tilde y_t^j), \quad \tilde u_t^{ij}=g^i_q(q_t, \tilde y_t^j).\] 
    \item Construct the affine representation
    \[
        \zeta^j = \mathcal{T}_0 w^j+ \sum_i \theta_i \mathcal{T}_1 \tilde u^{ij}.
    \]
    \item Solve the convex optimization problem:
    \[
        \theta^\star= \arg\min_\theta\;J(\zeta )+R(\theta) \quad \textrm{s.t.}\quad c(\zeta ^j)\le 0
    \]
where $R(\theta)$ is an optional regularization term.
\end{enumerate}

The result will of course only be approximately optimal, since $w^j$ are but a representative sample and the echo state network provides only a finite-dimensional span of policies. However it will be \textit{guaranteed} to be stabilizing.

\begin{remark}
This framework can be extended to include learning over all REN parameters, however the optimization problem is no longer convex. We have recently shown that this amounts to learning over all stabilizing nonlinear controllers for a linear system \cite{wang2022learning} and extended the framework to learn robustly stabilizing controllers for uncertain systems \cite{wang2021youla}.
\end{remark}

\subsection{Example}

We illustrate the approach on a simple discrete-time linear system with transfer function
\[
    \mathcal{T}_0=\mathcal{T}_1=-\mathcal{T}_2=\frac{0.3}{q^2-2\rho\cos(\phi)q+\rho^2}
\]
with $q$ the shift operator, $\rho = 0.8$, and $\phi = 0.2\pi$. We consider the task of minimizing the $\ell^1$ norm of the output in response to step disturbances, while keeping the control signal $u$ bounded: $|u_t|\le 5$ for all $t$. This can be considered a data-driven approach to an explicit model predictive control \cite{alessioSurveyExplicitModel2009} with stability guarantees.


Training data is generated by a 25,000 sample piece-wise constant disturbance that has a hold time of 50 samples and a magnitude uniformly distributed in the interval [-10, 10].

We construct a contracting model $\Qb$ with $n=50$ states and $q=500$ neurons by randomly sampling a matrix $X\in\R^{(2n+q)\times (2n+q)}$ with $X_{ij}\sim\mathcal{N}\left[0, \frac{4}{2n+q}\right]$ and constructing a C-REN via the method outline in Section \ref{sec:direct_caren}. The remaining parameters are sampled from the Glorot normal distribution \cite{glorot2010understanding}. For comparison, we construct a linear $\Qb$ parameter of the form
\begin{equation*}
	q_{t+1} = A_q q_t + B_q \tilde{y}_t, \quad v_{t+1} = C_q q_t + D_q \tilde{y}_t,
\end{equation*} 
where $A_q = \lambda\frac{\bar{A}}{\rho(\bar{A})}$ with $\lambda \in (0,1)$ and $\bar{A}_{ij} \sim \mathcal{N}\left[0, \frac{1}{2n + q}\right]$. Note that $A_q$ is a stable matrix with a contraction rate of $\lambda$. We sample $B_q$ from the Glorot normal distribution \cite{glorot2010understanding}.

\begin{figure}[]
	\centering
  \includegraphics[width=0.8\columnwidth]{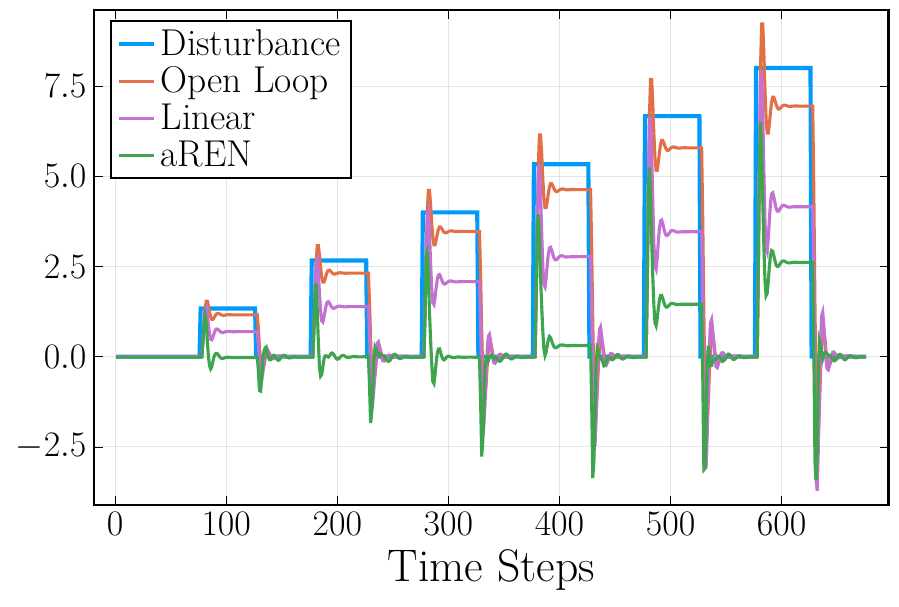}
  \includegraphics[width=0.8\columnwidth]{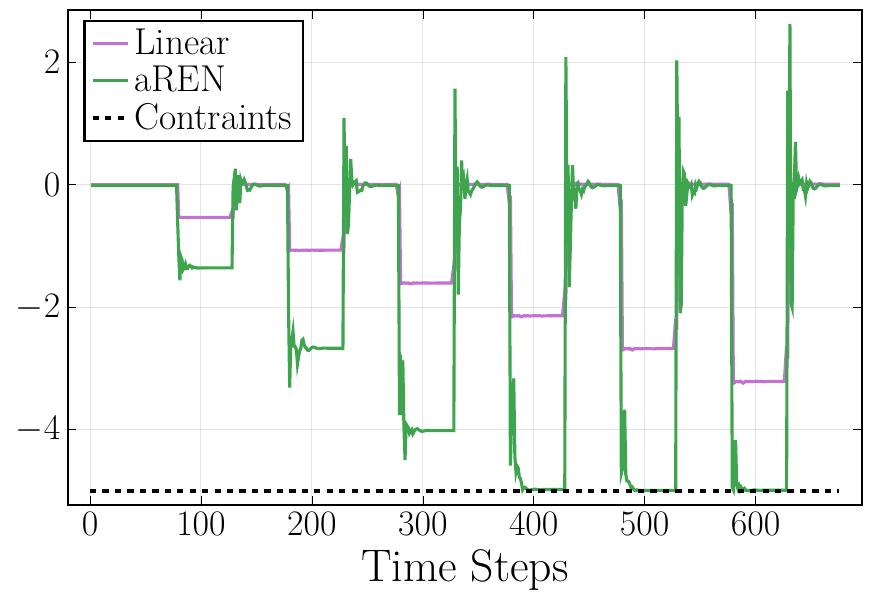}
  \caption{Output (top) and control signal (bottom) responses to step disturbances for nonlinear (C-REN) and linear  data-driven optimization of feedback controllers.}
  \label{fig:youla}
\end{figure}

The response to test inputs are shown in Fig. \ref{fig:youla}. The benefits of learning a nonlinear $\Qb$ parameter are that the control can respond aggressively to small disturbances, driving the output quickly to zero, but respond less aggressively to large disturbances to stay within the control bounds. In contrast, the linear control policy must respond proportionally to disturbances of all sizes. Since the control constraints require less aggressive response to large disturbances, the linear controller must also less aggressively to small disturbances, does not drive the output to zero.

\section{Conclusions}
In this paper we have \HL{introduced recurrent equilibrium networks (RENs) as a new model class} for learning nonlinear dynamical systems with built-in stability and robustness constraints. The model set is flexible and admits a direct parameterization, \HL{allowing learning of large-scale models via generic unconstrained optimization methods such as stochastic gradient descent}.

We have illustrated the benefits of the new model class on problems in system identification, observer design, and control. On system identification benchmarks, the REN structure outperformed the widely-used RNN and LSTM models in terms of model fit while achieving far lower sensitivity to input perturbations. We further showed that the REN model architecture enables new approaches to nonlinear observer design and optimization of nonlinear feedback controllers.

\appendix
\subsection{Proof of Theorem~\ref{prop:1}}\label{sec:proof-prop-1}
\Rev{
Firstly, well-posedness follows directly from \eqref{eq:lmi-stable-explicit}, since it implies $W\succ 0$ which is precisely \eqref{eq:W}.

To prove contraction and incremental IQCs we consider the incremental dynamics, i.e. differences between two sequences $(x^a,w^a,v^a,u^a)$ and $(x^b,w^b,v^b,u^b)$, which we denote $\Delta x_t = x_t^a-x_t^b$ and similarly for other variables.  The incremental dynamics  generated by \eqref{eq:rnn} are
\begin{align}
\begin{bmatrix}
\Delta x_{t+1} \\ \Delta v_t \\ \Delta y_t
\end{bmatrix}&=\begin{bmatrix}
A & B_1  & B_2\\
C_{1} & D_{11} & D_{12} \\
C_2 & D_{21} & D_{22} 
\end{bmatrix}\begin{bmatrix}
\Delta x_t \\ \Delta w_t \\ \Delta u_t
\end{bmatrix}, \label{eq:delta-G}\\
\Delta w_t &=\sigma(v_t^b+\Delta v_t)-\sigma(v_t^b). \label{eq:delta-sigma}
\end{align}

To deal with the nonlinear element \eqref{eq:delta-sigma}, we note that the constraint \eqref{eq:slope} can be rewritten as $(\sigma(x)-\sigma(y))(x-y)\ge (\sigma(x)-\sigma(y))^2$, and by taking a conic combinations of this inequality for each channel with multipliers $\lambda_i>0$, we obtain the following incremental quadratic constraint:
\begin{equation}\label{eq:iqc}
\Gamma(\Delta v, \Delta w)=
\begin{bmatrix}
\Delta v \\ \Delta w
\end{bmatrix}^\top
\begin{bmatrix}
0 & \Lambda \\
\Lambda & -2\Lambda
\end{bmatrix}
\begin{bmatrix}
\Delta v \\ \Delta w
\end{bmatrix}\geq 0,
\end{equation}
which is valid for any $\Lambda=\mathrm{diag}(\lambda_1,\ldots,\lambda_q)\in \Db_+$. 

To prove contraction, we first note that if \eqref{eq:lmi-stable-explicit} holds then
\begin{equation}\label{eq:lmi-stable-nonstrict}
		\begin{bmatrix}
			\alpha^2 P & -C_1^\top \Lambda \\
			-\Lambda C_1 & W
		\end{bmatrix}-
		\begin{bmatrix}
			A^\top \\ B_1^\top
		\end{bmatrix}P
		\begin{bmatrix}
			A^\top \\ B_1^\top
		\end{bmatrix}^\top\succeq 0
	\end{equation}
for some $\alpha<\bar \alpha$. Left-multiplying by $ \bigl[\Delta x_t^\top \; \Delta w_t^\top \bigr] $ and right-multiplying by $ \bigl[\Delta x_t^\top \; \Delta w_t^\top \bigr]^\top $, we obtain the following incremental Lyapunov inequality:\begin{equation}
	|\Delta x_{t+1}|^2_P \leq \alpha^2|\Delta x_t|_P^2-\Gamma(\Delta v_t, \Delta w_t) 
 \leq \alpha^2|\Delta x_t|_P^2.\label{eq:contraction_ineq}
\end{equation}
where the second inequality follows by the incremental quadratic constraint \eqref{eq:iqc}. Iterating over $t$ gives \eqref{eq:contraction} with $K=\sqrt{\bar\sigma/\underline{\sigma}}$
where $\bar\sigma$ is the maximum singular value of $P$, and $\underline{\sigma}$ the minimum singular value.
}

The proof for the incremental IQC is similar: from \eqref{eq:lmi-qsr-explicit} we obtain a non-strict version with $\alpha<\bar\alpha$. Left multiplying by $ \bigl[\Delta x_t^\top \; \Delta w_t^\top \Delta u_t^\top \bigr]$ and right-multiplying by its transpose results in:
\begin{align}
	|\Delta x_{t+1}|^2_P \leq & ~\alpha^2|\Delta x_t|_P^2-\Gamma(\Delta v_t, \Delta w_t) \notag \\&+\begin{bmatrix}
		\Delta y_t\\ \Delta u_t
	\end{bmatrix}^\top
	\begin{bmatrix}
		Q & S^\top \\ S & R
	\end{bmatrix}
	\begin{bmatrix}
		\Delta y_t \\ \Delta u_t
	\end{bmatrix}.\label{eq:diss_ineq1}
\end{align}
Since $\Gamma(\Delta v_t, \Delta w_t)\ge 0$ from \eqref{eq:iqc}, and $\alpha<1$ we have 
\begin{align}
	|\Delta x_{t+1}|^2_P -|\Delta x_t|_P^2\leq\begin{bmatrix}
		\Delta y_t\\ \Delta u_t
	\end{bmatrix}^\top
	\begin{bmatrix}
		Q & S^\top \\ S & R
	\end{bmatrix}
	\begin{bmatrix}
		\Delta y_t \\ \Delta u_t
	\end{bmatrix}.\notag
\end{align}
\Rev{Telescoping sum of the above inequality yields the IQC \eqref{eq:iqc-def} with $d(a,b)=(b-a)^\top P (b-a)$.  Moreover, since $Q\preceq 0$, taking $\Delta u_t=0$ in \eqref{eq:diss_ineq1} reduces to \eqref{eq:contraction_ineq} proving contraction.}

\subsection{Proof of Theorem~\ref{thm:2}}\label{sec:proof-thm-2}

We note that a REN has Lipschitz bound of $\gamma$ if \eqref{eq:lmi-qsr-2} holds with $Q=-\frac{1}{\gamma}I, R=\gamma I, S=0$. By taking Schur complements and permuting the third and fourth columns and rows, the condition to be verified can be rewritten as:
\begin{align}\label{eq:block matrix l2}
\renewcommand{\arraystretch}{1.1}
  \left[\begin{array}{ccc|cc}\bar\alpha^2 P & -C_1^\top \Lambda & A^\top&0& C_2^\top\\
				-\Lambda C_1 & W &B_1^\top& - \Lambda D_{12} &D_{21}^\top\\
				A & B_1  & P^{-1}&B_2&D_{22}^\top \\\hline
				0 &-D_{12}^\top\Lambda&B_2^\top&\gamma I&0\\
				 C_2&D_{21}&D_{22}& 0&
				 \gamma I\end{array}\right]\succ 0.
\end{align}
Now, the upper-left quadrant is positive-definite via Schur complement of \eqref{eq:lmi-stable-explicit}. Hence, by taking $\gamma$ sufficiently large, the condition \eqref{eq:block matrix l2} will be verified.

\subsection{Proof of Theorem \ref{thm:convex_param}}\label{sec:proof-thm-1}

To show well-posedness, from \eqref{eq:lmi-stable} we have $ E+E^\top \succ P \succ 0$ and $ \mathcal{W}=2\Lambda -\Lambda  {D_{11}}-{D_{11}^\top}\Lambda \succ 0 $  where $D_{11}=\Lambda^{-1}\mathcal{D}_{11}$. The first inequality implies that $E$ is invertible  and thus \eqref{eq:G} is well-posed. The second one ensures that the equilibrium network \eqref{eq:implicit} is well-posed by the main result of \cite{revay2020lipschitz}.

To prove contraction, applying the inequality $ \bar\alpha^2 E^\top \mathcal{P}^{-1}E\succeq E+E^\top-\tfrac{1}{\bar{\alpha}^2}\mathcal{P}$  \Rev{\cite[Sec. II]{tobenkinConvexParameterizationsFidelity2017}} and a Schur complement to \eqref{eq:lmi-stable} gives
\[
	\begin{bmatrix}
		\bar\alpha^2 E^\top \mathcal{P}^{-1}E & -\mathcal{C}_1^\top \\
		-\mathcal{C}_1 & \mathcal{W}
	\end{bmatrix}-
	\begin{bmatrix}
		F^\top \\ \mathcal{B}_1^\top
	\end{bmatrix}\mathcal{P}^{-1}
	\begin{bmatrix}
		F^\top \\ \mathcal{B}_1^\top
	\end{bmatrix}^\top \succ 0.
\]
By substituting $ F=EA$, $\mathcal{B}_1=EB_1$, $\mathcal{B}_2=EB_2$, $\mathcal{C}_1=\Lambda C_1$ and $\mathcal{D}_{11}=\Lambda D_{11}$ into the above inequality, we obtain \eqref{eq:lmi-stable-explicit} with $P=E^\top \mathcal{P}^{-1}E$. Thus, $\Theta_C$ is a set of C-RENs.
Similarly, we can show that \eqref{eq:lmi-qsr} implies \eqref{eq:lmi-qsr-explicit} for R-RENs.

\subsection{Proof of Proposition~\ref{prop:D22}}\label{sec:proof-prop-D22}
With the factorization $\mathcal Q = -L_Q^\top L_Q$, \eqref{eq:qsr-revised} is equivalent to 
\[
R-S\mathcal{Q}^{-1}S^\top \succ (L_QD_{22}-L_Q^{-\top}S^\top)^\top (L_QD_{22}-L_Q^{-\top}S^\top),
\]
which implies that $R-S\mathcal{Q}^{-1}S^\top \succ 0$, hence $L_R$ is well-defined.

If $p\geq m$, from \eqref{eq:M} we have $N^\top N \prec I$ since
\[
\begin{split}
    (I+&M)^\top(I+M)- (I+M)^\top N^\top N(I+M)\\
    &=2(M^\top+M)-4Z_3^\top Z_3=4(X_3^\top X_3+\epsilon I)\succ 0.
\end{split}
\]
Similarly, for the case $p<m$ we can obtain $N N^\top \prec I $ from \eqref{eq:M2}, which also implies $N^\top N \prec I$. Finally, by substituting \eqref{eq:D22} into \eqref{eq:qsr-revised} we have
\[
\begin{split}
    R+SD_{22}+D_{22}^\top S^\top +D_{22}^\top \mathcal{Q} D_{22} 
    =&L_R^\top(I-N^\top N)L_R\succ 0.
\end{split}
\]

\subsection{Proof of Proposition~\ref{prop:2}}\label{sec:proof-prop-2}

When the correctness condition \eqref{eq:correctness} holds, we have that $\hat x_t=x_t$ for all $t\ge 0$ if $\hat x_0=x_0$, i.e. the true state trajectory is a particular solution of the observer. But contraction implies that all solutions of the observer converge to each other. Hence when $w=0$ we have $\hat{x}_t\rightarrow x_t$ as $t\rightarrow \infty$.

Now we consider the case where the correctness condition does not hold but its error is bounded by \eqref{eq:obs_fit_err}. The dynamics of $\Delta x:=\hat{x}-x$ can be written as
\begin{equation*}
  \begin{split}
    \Delta x_{t+1}&=f_o(\hat{x}_t,u_t,y_t) - f_m(x_t,u_t) \\
    &=f_o(x_t+\Delta x_t,u_t,y_t)-f_o(x_t,u_t,y_t)+e_t
  \end{split}
\end{equation*}
where $e_t= f_o(x_t,u_t,y_t)-f_m(x_t,u_t)$.
By the mean-value theorem, $\Delta x_{t+1} = F(z,u_t)\Delta x_t + e_t$  where $F_t = \tfrac{\partial[f_o]}{\partial x}(z,u_t)$ for some $z$. By the triangle inequality
$
    |\Delta x_{t+1}|_P   \le  |F_t\Delta x_t|_P + |e_t|_P
$
and by contraction $ |F_t\Delta x_t|_P \le \alpha |\Delta x_t|_P$. So we have
\begin{align*}
   |\Delta x_{t+1}|_P -  |\Delta x_t|_P& \le (\alpha-1) |\Delta x_t|_P+|e_t|_P,\\&\le (\alpha-1) |\Delta x_t|_P+\sqrt{\bar\sigma} \rho.  
\end{align*}
From which it follows that the set $|\Delta x_t|_P\le \frac{\sqrt{\bar\sigma} \rho}{1-\alpha}$ is forward-invariant and exponentially attractive, since $\alpha-1<1$. The claimed result then follows from $\sqrt{\underline\sigma}|\Delta x_t|\le |\Delta_x|_P$.

\bibliographystyle{IEEEtran}
\bibliography{REN,ref}

\end{document}